\documentclass[11pt]{article}

\usepackage[final]{acl}

\usepackage{times}
\usepackage{latexsym}

\usepackage[T1]{fontenc}

\usepackage[utf8]{inputenc}

\usepackage{microtype}

\usepackage{inconsolata}

\usepackage{graphicx}
\usepackage{algorithm}
\usepackage{graphicx}
\usepackage{amsmath} 
\usepackage{amssymb} 
\usepackage{booktabs}
\usepackage{multirow}
\usepackage{float} 
\usepackage{amsmath}
\usepackage{newfloat}
\usepackage{listings}
\usepackage{algorithm}
\usepackage{algpseudocode}
\usepackage{xcolor}
\usepackage{amssymb}
\usepackage{fontawesome5}
\usepackage{simpleicons}

\definecolor{hfcolor}{HTML}{FFCC4D}

\title{Retrieval as Generation: A Unified Framework with Self-Triggered \\Information Planning}

\author{
Bo Li\textsuperscript{\rm 1,2},
Mingda Wang\textsuperscript{\rm 3},
Gexiang Fang\textsuperscript{\rm 1},
Shikun Zhang\textsuperscript{\rm 1},
Wei Ye\textsuperscript{\rm 1}\thanks{Corresponding author}
\\
\textsuperscript{\rm 1} National Engineering Research Center for Software Engineering, Peking University \\
\textsuperscript{\rm 2}School of Computer Science, Peking University \\
\textsuperscript{\rm 3} School of Health Sciences and Biomedical Engineering, Hebei University of Technology\\
\texttt{deepblue.lb@gmail.com, wye@pku.edu.cn}\\
\faGithub\ \href{https://github.com/WisdomShell/GRIP}{WisdomShell/GRIP}
\faGlobe\ \href{https://wisdomshell.github.io/GRIP/}{GRIP Project}
{\color{hfcolor}\simpleicon{huggingface}}\
\href{https://huggingface.co/collections/WisdomShell/grip}{HuggingFace Model}
}

\begin{document}
\maketitle
\begin{abstract}
We revisit retrieval-augmented generation (RAG) by embedding retrieval control directly into generation. Instead of treating retrieval as an external intervention, we express retrieval decisions within token-level decoding, enabling end-to-end coordination without additional controllers or classifiers. Under the paradigm of Retrieval as Generation, we propose \textbf{GRIP} (\textbf{G}eneration-guided \textbf{R}etrieval with \textbf{I}nformation \textbf{P}lanning), a unified framework in which the model regulates retrieval behavior through control-token emission. Central to GRIP is \textit{Self-Triggered Information Planning}, which allows the model to decide when to retrieve, how to reformulate queries, and when to terminate, all within a single autoregressive trajectory. This design tightly couples retrieval and reasoning and supports dynamic multi-step inference with on-the-fly evidence integration. To supervise these behaviors, we construct a structured training set covering answerable, partially answerable, and multi-hop queries, each aligned with specific token patterns. Experiments on five QA benchmarks show that GRIP surpasses strong RAG baselines and is competitive with GPT-4o while using substantially fewer parameters. 
\end{abstract}

\section{Introduction}

Retrieval-Augmented Generation (RAG) improves LLMs by grounding generation on external evidence, boosting open-domain QA, fact verification, and multi-hop reasoning~\cite{gao2023retrieval,kandpal2023large,Xu2024UnsupervisedIR,Luo2024LandmarkEA,yang2025benchmarkingmultimodalragchartbased,zhou2026adaptive,DBLP:conf/aaai/LiXX26}. 
However, most RAG systems keep retrieval external and one-shot: they retrieve documents from the initial query and then generate from a fixed context~\cite{Yu2023ImprovingLM,Shi2023REPLUGRB,Wang2023Query2docQE}, an assumption that can fail when information needs emerge gradually during step-by-step reasoning or when queries contain ambiguity and hidden dependencies.


To address this limitation, prior work explores retrieval control to improve timing and evidence use. 
Training-free methods (e.g., DRAGIN~\cite{su2024dragin}, FLARE~\cite{jiang2023active}) trigger retrieval via heuristic uncertainty signals, while training-based approaches learn retrieval-related behaviors with supervision, spanning both single-step control (e.g., INFO-RAG~\cite{Xu2024UnsupervisedIR}, Self-RAG~\cite{asai2024self}, GainRAG~\cite{jiang2025gainrag}) and agentic deep-search systems that iteratively generate queries (e.g., R1-Searcher~\cite{Song2025R1SearcherIT}, ZeroSearch~\cite{Sun2025ZeroSearchIT}). 
However, many of these methods still rely on auxiliary controllers or multi-stage procedures outside token-level generation, so retrieval timing, query reformulation, and stopping are not represented as explicit, trainable actions within a single decoding trajectory. 
This separation can also make it harder to attribute errors to specific decisions (e.g., retrieving too early vs.\ stopping too late) and to learn a consistent policy that couples retrieval with the model's evolving intermediate states.


Rather than relying on external modules to make discrete retrieval decisions, we embed retrieval behavior directly into the model's generative policy. Under the paradigm of Retrieval as Generation, we propose \textbf{GRIP} (\textbf{G}eneration-guided \textbf{R}etrieval with \textbf{I}nformation \textbf{P}lanning), a unified framework where retrieval is controlled by the same token-level decoding process as language generation\footnote{Here, ‘Retrieval as Generation’ does not mean that the retriever is internalized into the LLM. The retriever remains external; GRIP generates retrieval-control actions (e.g., retrieval triggering, query reformulation, and termination decisions) as part of the decoding trajectory, and the retrieved documents are then supplied back as context.}. GRIP regulates retrieval via explicit control tokens, including \texttt{[RETRIEVE]}, \texttt{[ANSWER]}, \texttt{[INTERMEDIARY]}, and \texttt{[SOLVED]}.  At its core, self-triggered information planning allows the model to decide when to retrieve, formulate follow-up queries conditioned on the evolving reasoning context, and determine when to stop and finalize the answer, all within a single autoregressive trajectory.

To support these behaviors, we introduce a structured training paradigm that supervises four answerability types with distinct control-token patterns. This supervision teaches the model to judge information sufficiency, trigger retrieval when needed, and terminate once the question is resolved. Unlike heuristic triggering, GRIP learns retrieval as a token-driven and context-sensitive capability. By integrating retrieval planning into generation, GRIP treats retrieval as an integral part of reasoning, enabling multi-step inference, adaptive retrieval depth, and robust generalization across tasks.

We conduct experiments on five knowledge-intensive QA benchmarks using multiple metrics. GRIP outperforms strong baselines and is competitive with proprietary API models. We further provide behavioral analyses of retrieval timing, query generation quality, and termination control, complemented by case studies and visualizations. In summary, our main contributions are as follows:

\begin{itemize}
    \item We propose \textbf{GRIP}, a unified \emph{Retrieval as Generation} framework that internalizes retrieval into token-level decoding via explicit control tokens, enabling retrieval planning, query reformulation, and termination within a single autoregressive trajectory.

    \item We extensively evaluate GRIP on five QA benchmarks and provide a mechanistic analysis of retrieval-as-generation behaviors. Beyond consistent gains over strong baselines, our studies show that GRIP learns task-aware retrieval depth and reliable stopping, yielding more adaptive and controllable retrieval trajectories across datasets.

\end{itemize}

\section{Generation-guided Retrieval with Information Planning}



\subsection{Token-Level Control for Unified Retrieval and Generation}
\label{sec:token_control}

We propose \textbf{GRIP}, a \textit{Retrieval-as-Generation} framework that internalizes retrieval control into token-level decoding. By emitting a small set of control tokens, the model decides when to retrieve, how to reformulate queries, and when to terminate. At the core is \textit{self-triggered information planning}, which assesses information sufficiency and executes multi-step retrieval. To realize this control in a unified and interpretable manner, GRIP augments the model’s output space with a small set of control tokens, allowing retrieval and termination decisions to be made directly during decoding.

GRIP introduces four tokens: \texttt{[RETRIEVE]} (request external evidence), \texttt{[INTERMEDIARY]} (emit a partial state), \texttt{[ANSWER]} (start the final response), and \texttt{[SOLVED]} (terminate). They form structured patterns that define two main branches: \texttt{[INTERMEDIARY]}$\rightarrow$\texttt{[RETRIEVE]} for continuing evidence acquisition, and \texttt{[ANSWER]}$\rightarrow$\texttt{[SOLVED]} for completion, enabling multi-step and recursive retrieval when needed.

This minimal interface keeps control within autoregressive generation, enabling end-to-end learning without external controllers.

\begin{figure*}[t]
    \centering
    \includegraphics[width=0.89\linewidth]{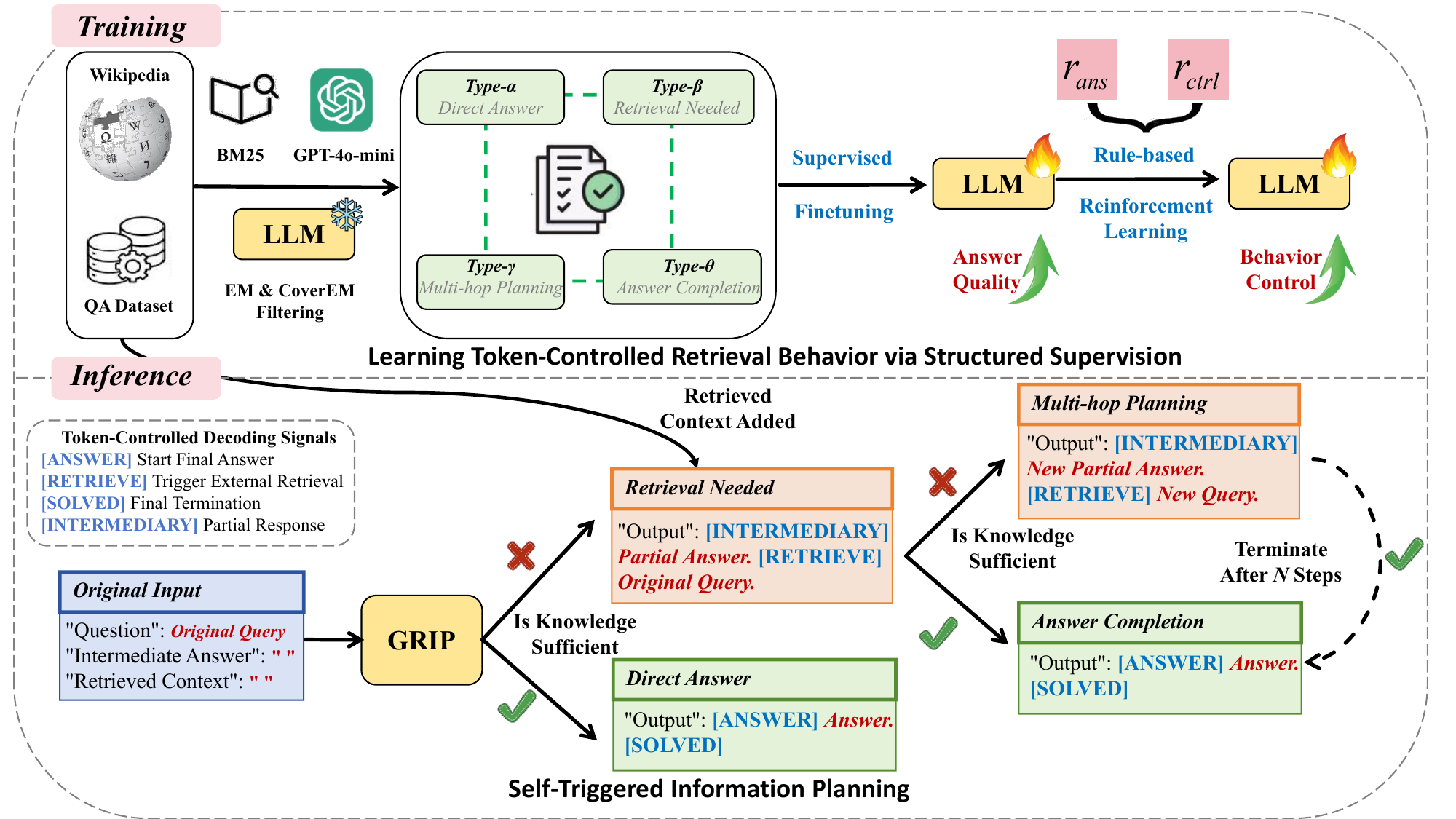}
    \caption{Overview of the GRIP framework. The top half illustrates the training process: four behavior types are constructed for supervised fine-tuning, followed by reinforcement learning to improve answer quality and retrieval control. The bottom half depicts self-triggered information planning, where GRIP dynamically emits control tokens to decide whether to retrieve, generate new queries, or terminate reasoning, all within a unified decoding trajectory.}
    \label{pic.model}
\end{figure*}

\subsection{Self-Triggered Information Planning}

GRIP adopts a language-native mechanism called \textit{self-triggered information planning}, which enables the model to autonomously decide when and how to retrieve information during generation. This process unfolds autoregressively, driven by the emission of control tokens. GRIP organizes its planning behavior into the following loop:

\paragraph{1. Initial Decision:} Given an original query, the model first determines whether its internal knowledge suffices. 
\begin{itemize}
    \item If so, it emits \texttt{[ANSWER]} followed by a final prediction, and then terminates with \texttt{[SOLVED]}.  
    \item If not, it emits \texttt{[INTERMEDIARY]} to provide partial knowledge or reasoning, followed by \texttt{[RETRIEVE]} and generates a new query for external retrieval.
\end{itemize}

\paragraph{2. Retrieval and New Query Generation:} The model uses the original or generated query to retrieve relevant passages, which are incorporated as augmented input. It then reassesses whether the available information is sufficient:

\begin{itemize}
    \item If the evidence is sufficient, the model emits \texttt{[ANSWER]} followed by a final prediction and concludes with \texttt{[SOLVED]}.
    \item Otherwise, it generates \texttt{[INTERMEDIARY]} explanation and emits a follow-up \texttt{[RETRIEVE]} token with a new query to continue the loop.
\end{itemize}

\textit{Notably, GRIP distinguishes between initial and follow-up retrievals:} 1) In the initial decision step, the \texttt{[RETRIEVE]} token is followed by the original query, as the model lacks any retrieved context. 2) In subsequent planning steps, the model conditions on its previously generated intermediate response and the original query to synthesize a new follow-up query tailored to unresolved aspects of the task.
    
\paragraph{3. Multi-hop Planning:} This alternation continues recursively: each \texttt{[INTERMEDIARY]} reflects the model’s current understanding, and each \texttt{[RETRIEVE]} introduces a targeted query to acquire additional knowledge.

\paragraph{4. Termination:} The retrieval-reasoning loop concludes when the model emits \texttt{[ANSWER]} followed by \texttt{[SOLVED]}, indicating that sufficient information has been gathered to produce a final answer. To ensure efficiency and behavioral stability, GRIP uses a decoding-time retrieval budget, which can be adjusted at inference time and is set to three retrieval rounds by default. If this limit is reached, the model is required to finalize its response: it emits \texttt{[ANSWER]} and generates the final answer based on the currently available information.

This planning mechanism is fully internal to the model and operates within a unified decoding flow. No confidence thresholds, external classifiers, or prompt chaining are needed. Through supervised training on structured examples, the model learns to coordinate retrieval and reasoning as a compositional skill. Pseudocode for the GRIP decoding process is provided in Appendix~\ref{app:code}.


\subsection{Learning Token-Controlled Retrieval Behavior via Structured Supervision}

\begin{figure}[t]
    \centering
    \includegraphics[width=1.0\linewidth]{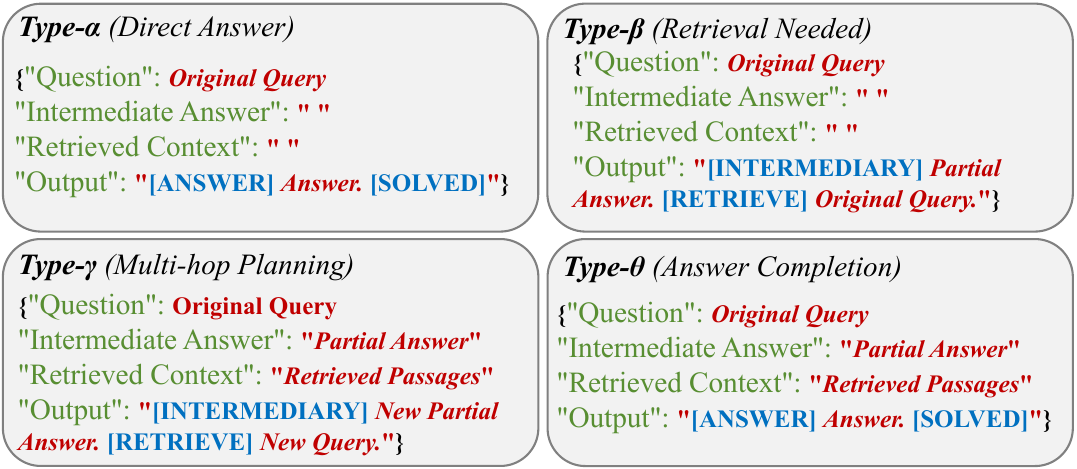}
    \caption{Four types of structured training samples in GRIP, each aligned with specific token patterns for controlling retrieval behavior. This design enables the model to learn when to retrieve, plan new queries, or finalize answers via language-native supervision.}
    \label{pic.data}
\end{figure}

To enable GRIP to learn retrieval as a generation-internal process, we formulate a structured supervision paradigm that aligns specific control token patterns with different retrieval behaviors. We organize training examples into four structured supervision scenarios, each reflecting a distinct behavioral trajectory. This token-controlled supervision allows GRIP to acquire compositional retrieval strategies, such as when to retrieve, how to formulate a new query, and when to terminate reasoning. We illustrate the four supervision types and their corresponding token structures in Figure~\ref{pic.data}.

\textbf{Type-\boldmath$\alpha$} samples represent directly answerable queries. These are queries that can be answered correctly using the model's internal knowledge alone, without any external context. We identify such instances by running the backbone model (e.g., LLaMA3-8B-instruct) and filtering cases that consistently yield exact-match (EM) answers across multiple decoding attempts. The model is trained to generate \texttt{[ANSWER]} followed by the final answer, and conclude with \texttt{[SOLVED]}.

\textbf{Type-\boldmath$\beta$} samples involve cases where the model provides partially correct or noisy answers that contain the gold answer but lack clarity. These cases are detected using a relaxed coverage-based EM metric, defined as responses that contain the target answer but fail to fully resolve the query. Here, the model is expected to emit \texttt{[INTERMEDIARY]} to reflect its current partial knowledge, followed by \texttt{[RETRIEVE]} and a retrieval query. This setup teaches the model to recognize uncertainty and trigger retrieval appropriately.

\textbf{Type-\boldmath$\gamma$} samples simulate multi-hop or complex knowledge cases where neither the model nor basic retrieval suffices. We identify such samples by detecting failures to answer correctly even with BM25-based retrieval, and then rely on GPT-4o-mini to generate an improved follow-up query focused on missing aspects. The model is trained to generate an intermediate answer (\texttt{[INTERMEDIARY]}) and a refined query (\texttt{[RETRIEVE]}), and to iterate through multiple planning steps before concluding with \texttt{[ANSWER]} and \texttt{[SOLVED]}.

\textbf{Type-\boldmath$\theta$} samples are cases where both the model and retriever can find passages that cover the answer superficially, but the correct response requires compressing, selecting, or synthesizing salient details. We use CoverEM to label these samples and train the model to refine retrieved content through reasoning. This teaches the model to emit token transitions like \texttt{[INTERMEDIARY]} $\rightarrow$ \texttt{[RETRIEVE]} $\rightarrow$ \texttt{[ANSWER]}, modeling selective information usage in noisy contexts.

Token supervision is applied using standard teacher forcing over these structured sequences. By associating each sample type with a precise token trajectory, the model learns to internalize retrieval control within its generative policy. Representative examples are provided in Appendix~\ref{app:data}.

\subsection{Training Objective and Optimization}~\label{sec:rl_reward}

GRIP is trained in two stages: (1) supervised fine-tuning on structured behavioral examples to learn retrieval patterns, and (2) rule-based reinforcement learning (RL) guided by two reward signals. Together, these stages enable the model to acquire retrieval planning skills that are both interpretable and adaptive.

\paragraph{Supervised Fine-Tuning.}  
The initial phase of GRIP training employs supervised learning to expose the model to diverse token-driven retrieval behaviors. Each training instance is labeled with a structured target sequence containing both reasoning content and control tokens (e.g., \texttt{[ANSWER]}, \texttt{[RETRIEVE]}), reflecting the semantic structure of the corresponding query type. The objective is to minimize the standard autoregressive cross-entropy loss over the entire target sequence:
\[
\mathcal{L}_{SFT} = -\sum_{t=1}^{T} \log P_\theta(y_t \mid y_{<t}, x)
\]
where \( x \) is the input query, and \( y_1, \dots, y_T \) is the ground-truth sequence containing both natural language and special control tokens. 

\paragraph{Reinforcement Fine-Tuning.}  
While SFT provides strong initial behavior modeling, we observe that the model tends to over-trigger retrieval actions, often emitting multiple \texttt{[RETRIEVE]} tokens even when unnecessary. To address this, we apply a RL phase based on rule-based reward optimization framework. We define two complementary reward signals:

\paragraph{1) Answer Fidelity Reward} \( r_{\text{ans}} \). Measures the similarity between the model's generated answer and the ground-truth reference using BLEU score~\cite{Chang2025BLEUBERIBI}, encouraging accurate and semantically faithful responses.

\paragraph{2) Control Accuracy Reward}\( r_{\text{ctrl}} \). Rewards the correct emission of control token patterns and penalizes incorrect or missing tokens, thereby sharpening the model’s understanding  of retrieval control logic. The final reward is computed as:
\[
R = r_{\text{ans}} + r_{\text{ctrl}}
\]
The model is optimized using DAPO~\cite{Yu2025DAPOAO}, encouraging answer-faithful and token-accurate behaviors that promote a more balanced retrieval strategy.

Overall, GRIP is designed to unify retrieval planning and language generation under a single autoregressive framework. Compared with closely related search-based methods, its advantage comes from two aspects: \textbf{(1) retrieval planning is internalized as token-level control within generation itself}, rather than implemented through external search-action loops; and \textbf{(2) multi-step retrieval behavior is learned via one-step decision optimization}, rather than long-horizon search-policy optimization. This makes GRIP both conceptually simple and practically stable, while still enabling adaptive retrieval depth and controllable stopping behavior at inference time.

\begin{table*}[h]
\small
\centering
\setlength{\tabcolsep}{0.55mm}
\renewcommand{\arraystretch}{1.07}
\begin{tabular}{ccccccccccccccccc}
\toprule[1.5pt]
\multicolumn{1}{c|}{\multirow{2}{*}{}}     & \multicolumn{3}{c|}{\textbf{HotpotQA}}                          & \multicolumn{3}{c|}{\textbf{PopQA}}                             & \multicolumn{3}{c|}{\textbf{NQ}}                                & \multicolumn{3}{c|}{\textbf{WebQ}}                             & \multicolumn{3}{c|}{\textbf{TriviaQA}}                          & \multirow{2}{*}{\textbf{\begin{tabular}[c]{@{}c@{}}Avg.\\ Score\end{tabular}}} \\ \cline{2-16}
\multicolumn{1}{c|}{}                      & \textbf{EM} & \textbf{ROUGE} & \multicolumn{1}{c|}{\textbf{F1}} & \textbf{EM} & \textbf{ROUGE} & \multicolumn{1}{c|}{\textbf{F1}} & \textbf{EM} & \textbf{ROUGE} & \multicolumn{1}{c|}{\textbf{F1}} & \textbf{EM} & \textbf{ROUGE} & \multicolumn{1}{c|}{\textbf{F1}} & \textbf{EM} & \textbf{ROUGE} & \multicolumn{1}{c|}{\textbf{F1}} &                                                                               \\ \toprule[1.5pt]
\multicolumn{17}{l}{\textit{\textbf{Training-free Method}}}                                                                                                                                                                                                                                                                                                                                                                                                          \\ \toprule[1.5pt]
\multicolumn{1}{c|}{\textbf{Instruct}}     & 17.2        & 21.6           & \multicolumn{1}{c|}{25.9}        & 17.4        & 19.6           & \multicolumn{1}{c|}{23.2}        & 14.4        & 23.8        & \multicolumn{1}{c|}{20.3}           & 14.8        & 25.1           & \multicolumn{1}{c|}{29.4}        & 46.2        & 45.3           & \multicolumn{1}{c|}{55.1}        & 26.6                                                                          \\
\multicolumn{1}{c|}{\textbf{GPT-3.5 Turbo}} & 26.2        & 32.4           & \multicolumn{1}{c|}{38.2}        & 29.1        & 31.1           & \multicolumn{1}{c|}{35.3}        & 20.7        & 34.3        & \multicolumn{1}{c|}{30.0}           & 15.9        & 27.4           & \multicolumn{1}{c|}{31.9}        & 55.6        & 57.8           & \multicolumn{1}{c|}{69.7}        & 35.7                                                                          \\
\multicolumn{1}{c|}{\textbf{GPT-4o}}       & 33.2        & 40.2           & \multicolumn{1}{c|}{47.0}        & 30.6        & 32.3           & \multicolumn{1}{c|}{39.9}        & 26.5        & 42.7        & \multicolumn{1}{c|}{28.3}           & 23.5        & 32.3           & \multicolumn{1}{c|}{37.0}        & 65.7        & 64.3           & \multicolumn{1}{c|}{78.2}        & \underline{41.4}                                                                    \\
\multicolumn{1}{c|}{\textbf{Single RAG}}   & 26.1        & 31.6           & \multicolumn{1}{c|}{37.2}        & 22.8        & 26.6           & \multicolumn{1}{c|}{31.1}        & 19.3        & 28.4        & \multicolumn{1}{c|}{24.8}           & 14.0        & 22.7           & \multicolumn{1}{c|}{26.6}        & 46.4        & 47.0           & \multicolumn{1}{c|}{56.8}        & 30.8                                                                          \\
\multicolumn{1}{c|}{\textbf{FLARE}}        & 23.2        & 27.9           & \multicolumn{1}{c|}{32.8}        & 14.3        & 16.0           & \multicolumn{1}{c|}{18.4}        & 14.7        & 22.4        & \multicolumn{1}{c|}{21.3}           & 24.2        & 30.3           & \multicolumn{1}{c|}{34.7}        & 48.6        & 48.5           & \multicolumn{1}{c|}{56.4}        & 28.9                                                                          \\
\multicolumn{1}{c|}{\textbf{DRAGIN}}       & 27.9        & 32.6           & \multicolumn{1}{c|}{38.7}        & 15.5        & 16.8           & \multicolumn{1}{c|}{19.8}        & 23.9        & 32.8        & \multicolumn{1}{c|}{28.5}           & 25.2        & 31.5           & \multicolumn{1}{c|}{35.7}        & 55.3        & 53.6           & \multicolumn{1}{c|}{64.6}        & 33.5                                                                          \\ 
\multicolumn{1}{c|}{\textbf{ETC}}       & 32.5        & 37.7           & \multicolumn{1}{c|}{44.2}        & 30.5        & 32.5           & \multicolumn{1}{c|}{37.5}        & 20.9        & 26.7        & \multicolumn{1}{c|}{30.7}           & 18.9        & 26.6           & \multicolumn{1}{c|}{30.4}        & 52.9        & 52.1           & \multicolumn{1}{c|}{63.0}        & 35.8                                                                          \\

\toprule[1.5pt]
\multicolumn{17}{l}{\textit{\textbf{Training-based Method}}}                                                                                                                                                                                                                                                                                                                                                                                                         \\ \toprule[1.5pt]
\multicolumn{1}{c|}{\textbf{SFT-RAG}}      & 20.3        & 24.1           & \multicolumn{1}{c|}{28.6}        & 29.4        & 25.2           & \multicolumn{1}{c|}{30.4}        & 20.8        & 17.9        & \multicolumn{1}{c|}{21.3}           & 18.9        & 18.3           & \multicolumn{1}{c|}{23.1}        & 50.1        & 24.8           & \multicolumn{1}{c|}{57.2}        & 27.4                                                                          \\
\multicolumn{1}{c|}{\textbf{Self-RAG}}     & 19.6        & 23.8           & \multicolumn{1}{c|}{26.7}        & 18.1        & 22.3           & \multicolumn{1}{c|}{22.8}        & 15.7        & 22.4        & \multicolumn{1}{c|}{24.0}           & 16.4        & 26.5           & \multicolumn{1}{c|}{27.4}        & 50.2        & 47.3           & \multicolumn{1}{c|}{57.5}        & 28.0                                                                          \\
\multicolumn{1}{c|}{\textbf{INFO-RAG}}     &  19.9        &    23.7            & \multicolumn{1}{c|}{26.9}            &     18.3        &    22.6            & \multicolumn{1}{c|}{23.0}            &    17.2         &    22.9         & \multicolumn{1}{c|}{24.9}               &    18.1         &    27.7            & \multicolumn{1}{c|}{28.9}            &    50.8         &    47.8            & \multicolumn{1}{c|}{58.1}            &    28.7                                                                           \\
\multicolumn{1}{c|}{\textbf{RobustRAG}}    & 27.6        & 31.8           & \multicolumn{1}{c|}{37.5}        & 29.7        & 27.7           & \multicolumn{1}{c|}{32.4}        & 26.4        & 25.1        & \multicolumn{1}{c|}{29.2}           & 21.5        & 25.0           & \multicolumn{1}{c|}{29.1}        & 48.8        & 47.7           & \multicolumn{1}{c|}{57.9}        & 33.2                                                                          \\
\multicolumn{1}{c|}{\textbf{GainRAG}}      & 31.4        & 35.6           & \multicolumn{1}{c|}{41.8}        & 30.1        & 33.3           & \multicolumn{1}{c|}{38.1}        & 22.9        & 27.9        & \multicolumn{1}{c|}{32.2}           & 16.5        & 24.5           & \multicolumn{1}{c|}{28.9}        & 50.3        & 49.1           & \multicolumn{1}{c|}{59.2}        & 34.8                                                                    \\
\multicolumn{1}{c|}{\textbf{R1-Searcher}}      & 26.0        & 29.1           & \multicolumn{1}{c|}{34.9}        & 41.6        & 35.2           & \multicolumn{1}{c|}{41.3}        & 25.8        & 24.9        & \multicolumn{1}{c|}{28.7}           & 21.8        & 26.1           & \multicolumn{1}{c|}{30.6}        & 56.0        & 53.3           & \multicolumn{1}{c|}{64.9}        & 36.0                                                                    \\
\multicolumn{1}{c|}{\textbf{RetRobust}}      & 29.6        & 34.9           & \multicolumn{1}{c|}{40.9}        & 34.1        & 35.1           & \multicolumn{1}{c|}{40.4}        & 24.2        & 29.0        & \multicolumn{1}{c|}{33.8}           & 21.8        & 27.4           & \multicolumn{1}{c|}{31.7}        & 53.6        & 50.9           & \multicolumn{1}{c|}{61.9}        & 36.6                                                                    \\
\multicolumn{1}{c|}{\textbf{InsturctRAG}}      & 31.2        & 36.8           & \multicolumn{1}{c|}{42.3}        & 33.1        & 35.7           & \multicolumn{1}{c|}{40.3}        & 29.5        & 29.5        & \multicolumn{1}{c|}{33.6}           & 19.5        & 26.8           & \multicolumn{1}{c|}{31.4}        & 51.3        & 52.2           & \multicolumn{1}{c|}{62.5}        & \underline{37.0}                                                                    \\
\multicolumn{1}{c|}{\textbf{GRIP}(ours)}         & 33.0        & 37.6           & \multicolumn{1}{c|}{44.1}        & 38.6        & 37.5           & \multicolumn{1}{c|}{38.4}        & 32.1        & 35.8        & \multicolumn{1}{c|}{32.0}           & 31.4        & 39.3           & \multicolumn{1}{c|}{34.6}        & 57.9        & 55.9           & \multicolumn{1}{c|}{67.4}        & \textbf{41.0\textsubscript{\scriptsize{+4.0}}}                                                                 \\
\multicolumn{1}{c|}{\textit{w/o RL}}       & 31.6        & 36.6           & \multicolumn{1}{c|}{43.0}        & 38.1        & 37.1           & \multicolumn{1}{c|}{37.6}        & 32.6        & 36.1        & \multicolumn{1}{c|}{32.7}           & 32.0        & 39.9           & \multicolumn{1}{c|}{35.1}        & 57.0        & 55.2           & \multicolumn{1}{c|}{66.8}        & 40.7                                                                          \\ \toprule[1.5pt]
\end{tabular}
\caption{
Main results on five QA benchmarks. We report EM, ROUGE (average of ROUGE-1/2/L), and F1; Avg.Score is the unweighted mean over all datasets and metrics. All \emph{open-source} baselines are reproduced using official implementations and released checkpoints under the same evaluation setting, and results are averaged over three random seeds. Results on a domain-specific dataset are provided in Appendix~\ref{app:domain}.}
\label{tab:main-results}
\end{table*}

\section{Experimental Setup and Main Results}
\subsection{Datasets and Evaluation Metrics}

\paragraph{Datasets.} We evaluate GRIP on five widely used benchmarks for open-domain and multi-hop question answering: \textbf{HotpotQA}~\cite{Ho2020ConstructingAM}, \textbf{PopQA}~\cite{Mallen2022WhenNT}, \textbf{Natural Questions (NQ)}~\cite{Kwiatkowski2019NaturalQA}, \textbf{WebQuestions (WebQ)}~\cite{Berant2013SemanticPO}, and \textbf{TriviaQA}~\cite{Joshi2017TriviaQAAL}.
 These datasets cover diverse query types, including factual recall, contextual understanding, and multi-step reasoning, providing a comprehensive testbed for retrieval-augmented generation. We use the official evaluation splits and verify that there is no overlap between evaluation questions and any training data used in our experiments. See Appendix~\ref{data0} for more detail.

\paragraph{Evaluation Metrics.}
We report three primary QA metrics: \textbf{Exact Match (EM)} (exact normalized string match), \textbf{ROUGE} (average of ROUGE-1/2/L), and \textbf{F1} (token-level F1 between prediction and reference). For compact reporting, we also use \textbf{Avg.Score}, defined as the unweighted mean of all reported metric values across all evaluated datasets (i.e., averaging over \emph{datasets} and \emph{metrics} with equal weight). We additionally report \textbf{CoverEM} for behavioral analysis; its definition and results are provided in Appendix~\ref{app:cover_em}.

\subsection{Experimental Details}

\paragraph{Structured Training Data Construction.} We construct structured training data based on the training sets of TriviaQA and NQ, each paired with three retrieved passages. The same top-3 retrieval setting is applied during inference for consistency. Specifically, we construct 40,000 structured training instances for the supervised fine-tuning stage, covering all four behavior types described in Section 2.3. For the RL stage, we generate an additional 5,000 samples with the same distribution across behavior types. The prompts used to construct structured data are provided in Appendix~\ref{app:prompt}. Additional settings for control tokens are described in Appendix~\ref{app:token}. Unless otherwise specified, we set the maximum retrieval steps to three, and we analyze larger test-time budgets in \S~\ref{sec:budget}.

\paragraph{Supervised Fine-tuning.} We fine-tune GRIP on 40,000 structured samples using LLaMA3-8B~\cite{grattafiori2024llama} as the backbone with full-parameter training. The model is trained for 8 epochs with a micro-batch size of 4 per GPU (total batch size 32 across 8 A800 GPUs). We use the AdamW with a learning rate of $1\times10^{-6}$. A cosine learning rate schedule is applied with 10\% warm-up. Input sequences are truncated at 8192 tokens. Due to space limitations, results using the Qwen2.5-7B-Instruct~\cite{qwen2.5,qwen3} are provided in the Appendix~\ref{app:qwen}.

\paragraph{Reinforcement Learning.}  
We adopt the DAPO~\cite{Yu2025DAPOAO} algorithm, an improved variant of GRPO~\cite{Shao2024DeepSeekMathPT}, to fine-tune GRIP on 5,000 structured samples. Two task-specific rewards are used: (1) \textit{Answer Fidelity}, computed via BLEU score against the reference answer, more details about the reward selection ablation can be found in the Appendix~\ref{app:reward_metric_ablation}; and (2) \textit{Control Accuracy}, which assigns 0.5 points for correctly emitted control tokens. We train the model for only 1 epoch with a learning rate of $1\times10^{-7}$, which is sufficient to stabilize retrieval behaviors under the DAPO framework. Please refer to Appendix~\ref{app:dapo_config} for more details.

\subsection{Comparison Models}

We compare GRIP with two groups of baselines. 
\textbf{Training-free} methods include direct prompting (\textbf{Instruct}), one-shot retrieval (\textbf{Single RAG}), proprietary API models (\textbf{GPT-3.5 Turbo}, \textbf{GPT-4o}) as references, and dynamic retrieval methods \textbf{FLARE}~\cite{jiang2023active}, \textbf{DRAGIN}~\cite{su2024dragin} and \textbf{ETC}~\cite{li2026modeling} (8-shot ICL following their original setups).
\textbf{Training-based} methods include a matched-data SFT baseline (\textbf{SFT-RAG}) and state-of-the-art frameworks \textbf{Self-RAG}~\cite{asai2024self}, \textbf{INFO-RAG}~\cite{Xu2024UnsupervisedIR}, \textbf{RobustRAG}~\cite{Fang2024EnhancingNR},\textbf{RetRobust}~\cite{yoran2024making}, \textbf{InstructRAG}~\cite{wei2024instructrag}, \textbf{RobustRAG}~\cite{Fang2024EnhancingNR}, \textbf{GainRAG}~\cite{jiang2025gainrag}, and \textbf{R1-Searcher}~\cite{Song2025R1SearcherIT}.

To ensure a fair comparison, we reproduce all \emph{open-source} baselines with official code/checkpoints when available and evaluate all models under the same protocol: LLaMA3-8B backbone, Wikipedia corpus, BM25 retriever with top-3 passages per query. For methods without explicit answer spans (e.g., INFO-RAG, RobustRAG, Self-RAG), we apply minimal output standardization for consistent scoring. We ablate retrievers (DPR and BM25+DPR) in Appendix~\ref{app:retriever_ablation} and use BM25 by default for its performance--efficiency trade-off.

\subsection{Main Results}

Table~\ref{tab:main-results} presents a comprehensive comparison between GRIP and a broad range of RAG methods. GRIP consistently achieves the best overall performance among open-source systems, outperforming strong training-based baselines such as GainRAG and RobustRAG across all five benchmarks, and also surpassing recent deep-search baselines such as R1-Searcher in Avg.Score. By internalizing retrieval decisions into the autoregressive generation process via token-level control, GRIP eliminates the need for external heuristics or controllers. Despite using a smaller backbone (LLaMA3-8B), GRIP reaches an Avg.Score of 41.0, close to GPT-4o, demonstrating strong effectiveness under a lightweight setting.

Gains are most pronounced on multi-hop or compositional benchmarks (e.g., HotpotQA and TriviaQA), where dynamic evidence acquisition is crucial. Beyond exact matching, GRIP also yields consistently higher ROUGE and F1, indicating better answer completeness and alignment. Overall, these results support GRIP as an efficient and generalizable framework for controlled retrieval-augmented generation across diverse QA scenarios.

\section{Analysis}

\subsection{Retrieval-Depth Adaptivity}

\begin{table}[h]
\small
\setlength{\tabcolsep}{0.65mm}
\renewcommand{\arraystretch}{1.1}
\begin{tabular}{c|ccc|ccc|c}
\toprule[1.5pt]
                 & \multicolumn{3}{c|}{\textbf{NQ}(54.3\%)}          & \multicolumn{3}{c|}{\textbf{WebQ}(24.3\%)}              & \multirow{2}{*}{\textbf{\begin{tabular}[c]{@{}c@{}}Avg.\\ Score\end{tabular}}} \\ \cline{2-7}
                 & \textbf{EM}    & \textbf{ROUGE}  & \textbf{F1} & \textbf{EM}    & \textbf{ROUGE}  & \textbf{F1} &                                     \\ \toprule[1.5pt]
\textbf{DRAGIN}  & 30.3          & 33.9          & 39.3          & 42.5          & 48.4          & 52.9          & 41.2                               \\
\textbf{GainRAG} & 31.6 & 34.4 & 39.6 & 33.0          & 40.7          & 45.6          & 37.4                               \\
\textbf{GRIP}    & 31.1          & 34.0          & 39.8          & 44.9 & 50.7 & 55.6 & \textbf{42.7}                      \\ \toprule[1.5pt]
\end{tabular}
\caption{Subset where DRAGIN/GainRAG retrieve at least once while GRIP answers without retrieval. Percentages denote the subset proportion per dataset.}

\label{tab:retrieval_elimination_comparison}
\end{table}

\begin{table}[h]
\small
\setlength{\tabcolsep}{0.55mm}
\renewcommand{\arraystretch}{1.07}
\begin{tabular}{c|ccc|ccc|c}
\toprule[1.5pt]
                 & \multicolumn{3}{c|}{\textbf{NQ}(23.1\%)}          & \multicolumn{3}{c|}{\textbf{WebQ}(43.8\%)}              & \multirow{2}{*}{\textbf{\begin{tabular}[c]{@{}c@{}}Avg.\\ Score\end{tabular}}} \\ \cline{2-7}
                 & \textbf{EM}    & \textbf{ROUGE}  & \textbf{F1} & \textbf{EM}    & \textbf{ROUGE}  & \textbf{F1} &                                     \\ \toprule[1.5pt]
\textbf{DRAGIN}  & 11.4            & 16.7          & 19.0          & 18.6          & 23.8          & 28.3          & 19.6                               \\
\textbf{GainRAG} & 8.8           & 14.8          & 17.4          & 11.1          & 18.4          & 22.4          & 15.5                                \\
\textbf{GRIP}    & 15.4          & 20.3          & 23.4          & 19.9          & 24.6          & 28.9          & \textbf{22.1}                      \\ \toprule[1.5pt]
\end{tabular}
\caption{Subset where GRIP performs two retrievals while DRAGIN/GainRAG retrieve at most once. Percentages denote the subset proportion per dataset.}
\label{tab:deep_retrieve_comparison}
\end{table}

A central benefit of token-level planning is \emph{bidirectional} control over retrieval depth: GRIP suppresses retrieval when the current decoding state is already sufficient, yet deepens retrieval when evidence is missing.
To make this behavior concrete, we study two complementary subsets.

First, we consider cases where DRAGIN/GainRAG perform at least one retrieval while GRIP answers directly; Table~\ref{tab:retrieval_elimination_comparison} shows that GRIP achieves higher accuracy on this subset despite using fewer retrieval calls, indicating it can avoid redundant retrieval rather than following a fixed schedule.

Second, we analyze cases where GRIP performs two retrievals while baselines retrieve at most once; Table~\ref{tab:deep_retrieve_comparison} shows consistent gains on these harder examples, suggesting GRIP can detect evidence insufficiency and extend retrieval accordingly.
Together, these results support that GRIP learns context-sensitive retrieval depth control from structured token supervision.

\subsection{Adaptive Retrieval Across Tasks}

Table~\ref{tab:tims} reports the mean number of retrieval calls per example. GRIP exhibits clear task-aware adaptivity in retrieval depth: it retrieves more frequently on HotpotQA (1.44) and PopQA (1.58), where multi-hop reasoning or long-tail entities often require external evidence~\cite{Mallen2022WhenNT,asai2024self}, while triggering substantially fewer retrievals on NQ (0.76), where many questions can be answered from the model’s parametric knowledge. Compared with training-based baselines that follow nearly fixed retrieval schedules (e.g., GainRAG always retrieves once), GRIP adjusts retrieval frequency across datasets without external controllers. Moreover, R1-Searcher performs markedly more retrievals (4.67--5.75 across datasets; 5.12 on average), indicating a substantially higher retrieval cost despite its weaker end performance in our setting. In contrast, GRIP achieves strong accuracy with a much lower retrieval budget.

\begin{table}[h]
\small
\centering
\setlength{\tabcolsep}{0.8mm}
\renewcommand{\arraystretch}{1.07}
\begin{tabular}{c|ccccc|c}
\toprule[1.5pt]
                    & \textbf{Hotpot} & \textbf{PopQA} & \textbf{NQ}    & \textbf{WebQ} & \textbf{Trivia} & {\textbf{\begin{tabular}[c]{@{}c@{}}Avg.\\ Count\end{tabular}}} \\ \toprule[1.5pt]
\textbf{DRAGIN}     & 1.13            & 1.26          & 1.06          & 1.02          & 1.08             & 1.11              \\
\textbf{GainRAG}    & 1.0             & 1.0          & 1.0          & 1.0          & 1.0             & 1.0              \\
\textbf{R1-Searcher}    & 5.75             & 5.34          & 5.09          & 4.67          & 4.76             & 5.12              \\
\textbf{GRIP}       & 1.44    & 1.58 & 0.76 & 1.15 & 1.25    & 1.24     \\ 
\textit{w/o RL}       & 1.99    & 1.98 & 0.77 &1.26  & 1.99    & 1.60     \\ 
\toprule[1.5pt]
\end{tabular}
\caption{Mean retrieval count per dataset. GRIP adapts retrieval frequency across tasks, and RL further reduces retrieval calls while preserving this adaptivity. Hotpot and Trivia denote HotpotQA and TriviaQA, respectively.}
\label{tab:tims}
\end{table}

Besides, rule-based RL further refines GRIP’s retrieval policy by consistently reducing redundant calls across \emph{all} datasets: the overall average drops from 1.60 to 1.24 (\(\sim\)22.5\% reduction), while preserving the same task-aware retrieval pattern. A finer-grained breakdown of this behavioral shift is provided in Appendix~\ref{app:shift}.


\subsection{Improving Retrieval Quality by Generating New Queries}

\begin{figure}[h]
    \centering
    \includegraphics[width=0.79\linewidth]{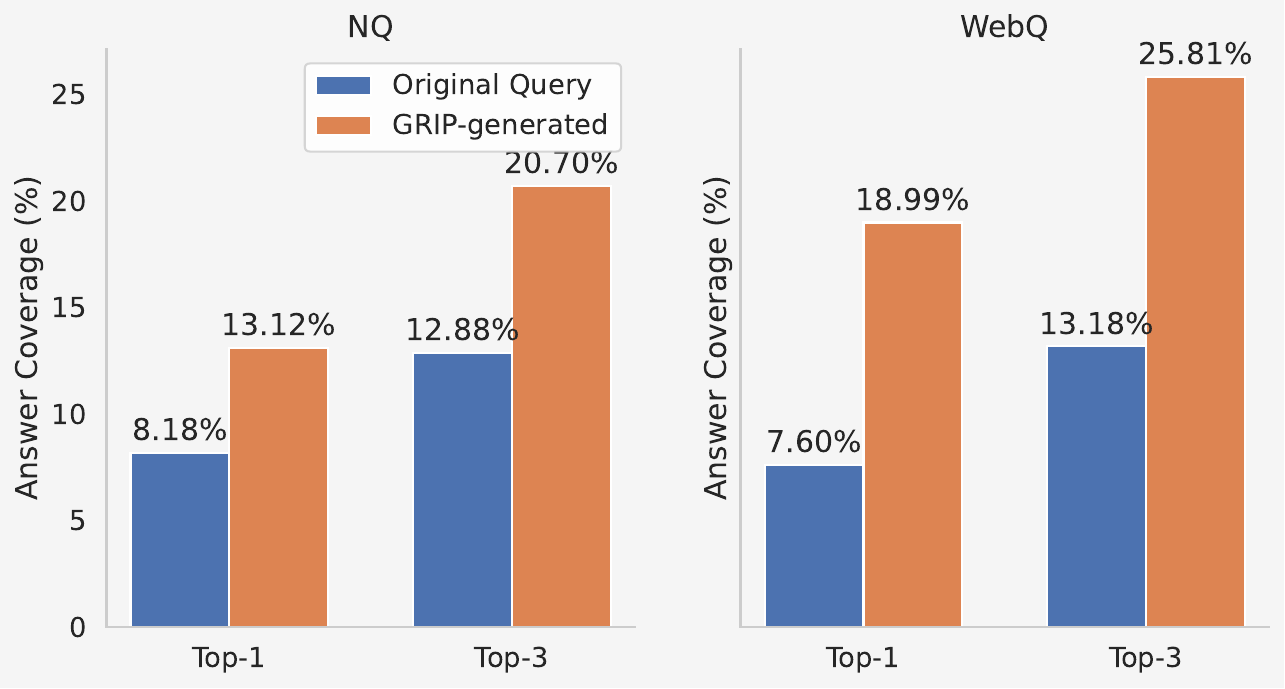}
    \caption{The percentage of samples where the gold answer span is found in the top-1 or top-3 retrieved passages. GRIP-generated queries significantly improve answer presence in retrieved contexts across both NQ and WebQ, indicating more effective retrieval behavior.}
    \label{pic.quality}
\end{figure}

To evaluate the effectiveness of GRIP’s self-triggered information planning, we analyze the retrieval quality of its generated new queries compared to the original ones. Specifically, we measure the proportion of top-1 and top-3 retrieved passages that contain the gold answer. As shown in Figure~\ref{pic.quality}, GRIP significantly improves retrieval coverage on both NQ and WebQ. This improvement does not stem from superficial query rewriting. Instead, it emerges from GRIP’s iterative reasoning process. By leveraging intermediate answers and previously retrieved content, GRIP gradually refines its focus toward unresolved aspects of the question. This progressive formulation yields more targeted new queries that are more likely to retrieve passages containing the correct answer. Notably, larger gains on top-1 than top-3 suggest that GRIP improves evidence ranking, consistent with query reformulation being coupled to intermediate reasoning states for more efficient retrieval.

\subsection{Controllable Retrieval Budget and Depth Extrapolation}\label{sec:budget}

We vary the maximum allowed retrieval steps $B$ at inference time and report both the realized retrieval frequency (Avg.Count) and overall Avg.Score (Table~\ref{tab:budget_control}). 
As $B$ increases from 3 to 10, Avg.Score rises from 41.0 to 41.8 while Avg.Count grows only from 1.24 to 1.62. 
Avg.Count stays far below the budget cap across all settings, indicating that GRIP does not blindly consume the available budget but retrieves adaptively when the decoding trajectory suggests an information gap, supporting our token-level planning claim.

\begin{table}[h]
\small
\centering
\setlength{\tabcolsep}{0.9mm}
\renewcommand{\arraystretch}{1.12}
\begin{tabular}{c|cccccc|c}
\toprule[1.5pt]
\multirow{2}{*}{\textbf{Max $B$}} & \multicolumn{6}{c|}{\textbf{Avg.Count (per dataset)}} & \multirow{2}{*}{\textbf{\begin{tabular}[c]{@{}c@{}}Avg.\\ Score\end{tabular}}}\\
& Hotpot & PopQA & NQ & WebQ & Trivia & Avg & \\
\toprule[1.5pt]
3  & 1.44 & 1.58 & 0.76 & 1.15 & 1.25 & 1.24 & 41.0\\
5  & 1.56 & 1.73 & 1.01 & 1.45 & 1.31 & 1.41 & 41.2\\
7  & 1.66 & 1.79 & 1.12 & 1.65 & 1.35 & 1.51 & 41.5\\
10 & 1.74 & 1.86 & 1.22 & 1.90 & 1.39 & 1.62 & 41.8\\
\bottomrule[1.5pt]
\end{tabular}
\caption{Effect of the maximum retrieval budget $B$ at inference time. Hotpot and Trivia denote HotpotQA and TriviaQA, respectively.}
\label{tab:budget_control}
\end{table}

This adaptive control also exhibits \emph{extrapolative generalization}: although training trajectories contain at most three retrieval steps, GRIP benefits from larger test-time budgets ($B{=}5,7,10$) with consistent gains, suggesting it learns when additional evidence is needed rather than a fixed-step pattern. 
Extra retrieval is mostly allocated to evidence-uncertain benchmarks (NQ: 0.76$\rightarrow$1.22; WebQ: 1.15$\rightarrow$1.90), and improvements show diminishing returns. Overall, $B$ provides a practical control parameter to trade accuracy for cost, while GRIP remains adaptive beyond the retrieval depth seen in training. Appendix~\ref{app:budget} shows the distribution of retrieval counts.

\subsection{Replacing Proprietary Teachers with Open-Source Supervision}
\label{sec:teacher_replacement}

\begin{table}[h]
\small
\centering
\setlength{\tabcolsep}{4pt}
\renewcommand{\arraystretch}{1.06}
\begin{tabular}{l l c}
\toprule[1.5pt]
\textbf{Teacher} & \textbf{Training} & \textbf{Avg.Score} \\
\toprule[1.5pt]
\multirow{2}{*}{GPT-4o-mini} 
  & GRIP (SFT)    & 40.7 \\
  & GRIP (SFT+RL) & \textbf{41.0} \\
\midrule
\multirow{2}{*}{LLaMA3-8B-Instruct}
  & GRIP (SFT)    & 38.8 \\
  & GRIP (SFT+RL) & 39.0 \\
\midrule
\multirow{2}{*}{Qwen3-32B}
  & GRIP (SFT)    & 40.2 \\
  & GRIP (SFT+RL) & 40.3 \\
\toprule[1.5pt]
\end{tabular}
\caption{Teacher replacement results, where only the GPT-labeled subset is re-annotated.}
\label{tab:teacher_replacement}
\end{table}

To test whether GRIP’s gains rely on proprietary distillation, we conduct a controlled teacher-replacement study on \textbf{Type-$\gamma$} supervision, which is the only place where our data construction uses an external teacher. Specifically, Type-$\gamma$ samples use GPT-4o-mini to generate improved follow-up retrieval queries when the backbone model fails. We replace GPT-4o-mini and re-annotate the follow-up queries for \emph{all} Type-$\gamma$ samples using two open teachers: LLaMA3-8B-Instruct and Qwen3-32B. All other components are kept identical across settings, including the construction and supervision of other samples, the prompts and decoding hyperparameters for annotation.

Table~\ref{tab:teacher_replacement} shows that GRIP remains strong under open teachers, indicating that the gains primarily arise from our framework design rather than GPT-specific artifacts. In particular, replacing GPT-4o-mini with Qwen3-32B yields a comparable final performance (40.3 vs.\ 41.0), suggesting that GRIP can still induce effective dynamic retrieval and query reformulation behaviors without relying on closed-source supervision. Using a backbone-scale teacher (LLaMA3-8B-Instruct) leads to a larger drop, implying that teacher capacity mainly affects the \emph{quality} of trajectory instantiation rather than the \emph{existence} of the learned retrieval policy. Overall, teacher outputs serve as interchangeable demonstrations, while the token-controlled supervision is the key factor that turns retrieval into a learnable generation skill.

\subsection{General Capability Preservation}
\label{sec:general_capability}

\begin{table}[h]
\centering
\small
\setlength{\tabcolsep}{8pt}
\begin{tabular}{lcc}
\toprule[1.5pt]
Model & MMLU (Acc) & MBPP (Pass@1) \\
\toprule[1.5pt]
Instruct & 66.56 & 54.6 \\
SFT-RAG & 62.93 & 47.8 \\
GRIP & 65.73 & 53.8 \\
\toprule[1.5pt]
\end{tabular}
\caption{General capability evaluation on closed-form non-RAG benchmarks.}
\label{tab:general_capability_closed}
\end{table}

A potential concern is whether fine-tuning GRIP for retrieval planning may degrade the model's general-purpose abilities on non-RAG tasks. To directly assess this, we conduct additional evaluations on two types of tasks that involve no retrieval: (1) closed-form knowledge and coding benchmarks, including MMLU~\cite{hendrycks2020measuring} and MBPP~\cite{austin2021program}, and (2) an open-ended generation task, CNN/DailyMail summarization~\cite{hermann2015teaching}, evaluated by GPT-4o as a pairwise judge.

\paragraph{Closed-form benchmarks.}
Table~\ref{tab:general_capability_closed} reports results on MMLU and MBPP. Compared with the base \textsc{Instruct} model, GRIP shows only a minor decrease, dropping by 0.83 points on MMLU and 0.8 points on MBPP. This suggests that learning retrieval-planning behaviors does not substantially harm the model's general knowledge, reasoning, or coding ability. In contrast, the matched fine-tuning baseline \textsc{SFT-RAG} exhibits noticeably larger degradation on both benchmarks, indicating that GRIP better preserves the backbone model's general capabilities while still introducing retrieval-oriented behaviors.

\paragraph{Open-ended summarization.}
We further evaluate open-ended generation quality on CNN/DailyMail summarization, where no retrieval is involved. Following recent practice, we use GPT-4o as a pairwise judge to compare model outputs. To reduce position bias, each model pair is evaluated in both orders, and we report the order-averaged Win/Equal/Loss rates in Table~\ref{tab:general_capability_open}. GRIP remains competitive with the base \textsc{Instruct} model, achieving a 51.0\% win rate versus 11.5\% loss rate, and is preferred much more often than other fine-tuned or search-oriented baselines such as \textsc{SFT-RAG} and \textsc{R1-Searcher}. These results suggest that GRIP preserves general-purpose generation quality while learning retrieval control.

\begin{table}[t]
\centering
\small
\setlength{\tabcolsep}{6pt}
\begin{tabular}{lccc}
\toprule[1.5pt]
Pairwise comparison (GRIP vs.) & Win & Equal & Loss \\
\toprule[1.5pt]
Instruct & 51.0 & 37.5 & 11.5 \\
SFT-RAG & 90.5 & 8.5 & 1.0 \\
R1-Searcher & 59.5 & 32.5 & 8.0 \\
\toprule[1.5pt]
\end{tabular}
\caption{Pairwise GPT-4o evaluation on CNN/DailyMail summarization. Results are averaged over both output orders to mitigate position bias.}
\label{tab:general_capability_open}
\end{table}

Overall, across both closed-form and open-ended non-RAG evaluations, GRIP shows minimal degradation relative to the base \textsc{Instruct} model, while substantially outperforming retrieval-oriented baselines in preserving general-purpose capabilities. This indicates that integrating retrieval planning into token-level generation does not come at the cost of broadly useful language modeling ability.

\section{Related Work}

Most standard RAG pipelines retrieve documents in a static, one-shot fashion based on the initial query~\cite{gao2023retrieval,fan2024survey,xiong2024search,ye2024r2ag,zhang2025less,zhang2026stable,DBLP:conf/aaai/LiXX26}. To reduce unnecessary retrieval, several works estimate retrieval necessity at inference time using heuristic uncertainty signals, such as entropy-based filtering or logit-margin thresholds~\cite{jiang2023active,ram2023context,su2024dragin,li2026modeling}. While lightweight, these approaches rely on hand-crafted criteria rather than learning retrieval behaviors end-to-end.

Recent studies make retrieval behavior learnable by introducing retrieval-relevant supervision, e.g., self-reflection/critique signals or contrastive/adversarial objectives~\cite{asai2024self,jiang2025gainrag,Xu2024UnsupervisedIR,Fang2024EnhancingNR}. In parallel, deep search and multi-step retrieval systems further strengthen search-and-reason routines for complex queries~\cite{Song2025R1SearcherIT,Sun2025ZeroSearchIT}, as well as planning-oriented multi-hop methods~\cite{Lee2024PlanRAGAP,Jin2025SearchR1TL,Yan2025RPORP,zhu2026symphony,xu2026selfcorrectingrag}. However, many of these approaches still rely on external controllers or multi-stage pipelines for retrieval decisions, which can complicate integration and reduce transparency. In contrast, GRIP internalizes retrieval as token-level generative actions, realizing timing, query reformulation, and termination within a single autoregressive trajectory.

\section{Conclusion}
We present GRIP, a unified retrieval-augmented generation framework that integrates retrieval behavior into the generation process through self-triggered information planning. This design enables dynamic control over retrieval without relying on external modules. Experimental results across five QA benchmarks confirm that GRIP consistently outperforms existing RAG methods and achieves performance comparable to GPT-4o. We hope this work provides a foundation for future efforts in controllable and efficient generation.

\section*{Limitations}
Although GRIP learns to regulate retrieval through discrete control tokens, its behavior remains sensitive to the design of the retrieval interface, such as the maximum retrieval budget $B$ and the evidence packing strategy (e.g., chunking and top-$k$). In this work we keep these choices fixed for fair comparison, but more adaptive evidence budgeting and context structuring could further improve robustness across domains.

\bibliography{custom}

\clearpage
\appendix
\section*{Appendix}
\section{Test Dataset Statistics}\label{data0}

\subsection{Dataset. }We evaluate GRIP on \textbf{five} QA benchmarks as the main evaluation suite in the main paper, and \textbf{additionally} report results on \textbf{BioASQ} in this appendix as a domain-specific benchmark. HotpotQA focuses on multi-hop reasoning. PopQA, NQ, and WebQ are open-domain QA datasets, targeting factual questions with varying entity popularity and coverage. TriviaQA emphasizes reading comprehension from web and Wikipedia sources, featuring complex, compositional questions authored by trivia enthusiasts. Finally, BioASQ contains biomedical questions that require specialized knowledge and is \emph{not} included in the main-suite average unless explicitly stated. These datasets collectively test the model’s adaptability to different retrieval and reasoning demands.

\begin{table}[h]
\small
\centering
\begin{tabular}{c|c|c}
\toprule[1.5pt]
                         & \textbf{\#Samples} & \textbf{Type}         \\ \toprule[1.5pt]
\textbf{HotpotQA}        & 7,405               & Multi-hop             \\
\textbf{PopQA}           & 14,267              & Open-domain           \\
\textbf{NQ} & 3,610               & Open-domain           \\
\textbf{WebQ}           & 2,032               & Open-domain           \\
\textbf{TriviaQA}       & 11,313              & Reading Comprehension \\
\textbf{BioASQ}          & 885                & Domain-specific       \\ \toprule[1.5pt]
\end{tabular}
\caption{Statistics and types of the test datasets used in our evaluation. The datasets vary in size and reasoning focus, covering multi-hop, open-domain, reading comprehension, and biomedical QA tasks.}
\end{table}

\subsection{Evaluation Metrics. }
We report three automatic metrics on all QA benchmarks, where higher is better.  For each test instance, we compute the metric against the corresponding reference answer(s), and then average scores over the evaluation set.

\paragraph{Exact Match (EM).}
EM is a binary metric that equals 1 if the predicted answer exactly matches any reference answer after standard normalization (e.g., lowercasing and trimming extra whitespaces), and 0 otherwise. The final EM is the average over all instances.

\paragraph{Token-level F1.}
F1 measures the token overlap between the prediction and a reference answer. 
Following the standard QA evaluation protocol, we compute precision and recall based on overlapping tokens, and the final F1 is averaged across instances.

\paragraph{ROUGE.}
We compute ROUGE-$n$ for $n\in\{1,2,L\}$ between the generated output and the reference text(s). 
We then report $\mathrm{ROUGE}=\frac{1}{3}(\mathrm{ROUGE}\text{-}1+\mathrm{ROUGE}\text{-}2+\mathrm{ROUGE}\text{-}L)$.

\section{Performance On The Domain-specific Dataset}\label{app:domain}

\begin{table}[h]
\small
\centering
\setlength{\tabcolsep}{3.6mm}
\renewcommand{\arraystretch}{1.08}
\begin{tabular}{c|cc|c}
\toprule[1.5pt]
 & \textbf{ROUGE} & \textbf{F1} & \textbf{Avg.Score} \\
\midrule
\textbf{FLARE}     & 30.2 & 69.4 & 49.8 \\
\textbf{DRAGIN}    & 47.1 & 81.2 & 64.2 \\
\textbf{SFT-RAG}   & 39.1 & 58.6 & 48.9 \\
\textbf{Self-RAG}  & 45.9 & 68.8 & 57.4 \\
\textbf{INFO-RAG}  & 39.4 & 59.2 & 49.3 \\
\textbf{RobustRAG} & 53.9 & 80.9 & 67.4 \\
\textbf{GainRAG}   & 49.3 & 74.0 & 61.7 \\
\textbf{GRIP}      & 54.8 & 84.4 & \textbf{69.6} \\
\bottomrule[1.5pt]
\end{tabular}
\caption{Performance on the domain-specific BioASQ benchmark (higher is better). Avg.Score is the unweighted mean of ROUGE and F1.}
\label{tab:domain}
\end{table}

We further evaluate all models on BioASQ, a domain-specific QA benchmark where questions often require specialized biomedical knowledge beyond the parametric capacity of general-purpose LLMs. As shown in Table~\ref{tab:domain}, GRIP achieves the best overall performance, obtaining the highest ROUGE (54.8) and F1 (84.4), and thus the highest Avg.Score (69.6). Compared with strong training-based baselines such as RobustRAG and GainRAG, GRIP yields more accurate and better-supported answers, suggesting that its self-triggered information planning can more reliably decide when to retrieve and how to formulate domain-relevant queries under high knowledge uncertainty.

\section{Cover Match Analysis for Answer Stability}\label{app:cover_em}

To further assess semantic correctness beyond exact matching, we report CoverEM scores across all datasets. CoverEM is designed to measure whether the model output covers the reference answer string. 
After applying the same text normalization as EM, CoverEM equals 1 if the normalized prediction contains the normalized reference answer string as a contiguous substring (for any reference in the reference set), and 0 otherwise. The final CoverEM is averaged across instances.

As shown in Table~\ref{tab:coverem_results}, GRIP significantly outperforms all baselines, including powerful models like GPT-4o and GainRAG. This suggests that even when predictions diverge from the gold answer in surface form, GRIP still generates semantically aligned spans with correct content. The consistent gains across datasets suggest that GRIP’s outputs are not only accurate but also robust to surface-level variations.

\begin{table}[h]
\small
\centering
\setlength{\tabcolsep}{0.3mm}
\renewcommand{\arraystretch}{1.07}
\begin{tabular}{c|ccccc|c}
\toprule[1.5pt]
                    & \textbf{Hotpot} & \textbf{PopQA} & \textbf{NQ}    & \textbf{WebQ} & \textbf{Trivia} & {\textbf{\begin{tabular}[c]{@{}c@{}}Avg.\\ CoverEM\end{tabular}}} \\ \toprule[1.5pt]
\textbf{GPT-4o}     & 38.6             & 45.2          & 48.5          & 43.6          & 82.1             & 51.6              \\
\textbf{DRAGIN}     & 31.6             & 23.0          & 38.4          & 41.5          & 68.0             & 40.5              \\
\textbf{GainRAG}    & 38.0             & 44.4          & 39.1          & 36.9          & 64.2             & 44.5              \\
\textbf{GRIP}       & 45.0    & 49.1 & 41.2 & 52.4 & 74.9    & \textbf{52.5}     \\ \toprule[1.5pt]
\end{tabular}
\caption{CoverEM scores across five QA datasets. GRIP achieves the best average CoverEM. Hotpot and Trivia denote HotpotQA and TriviaQA, respectively.}
\label{tab:coverem_results}
\end{table}

\section{Structured Dataset Samples}\label{app:data}
We now provide several dataset samples to help readers understand the structured training data easily.

\begin{figure}[h]
    \centering
    \includegraphics[width=1.0\linewidth]{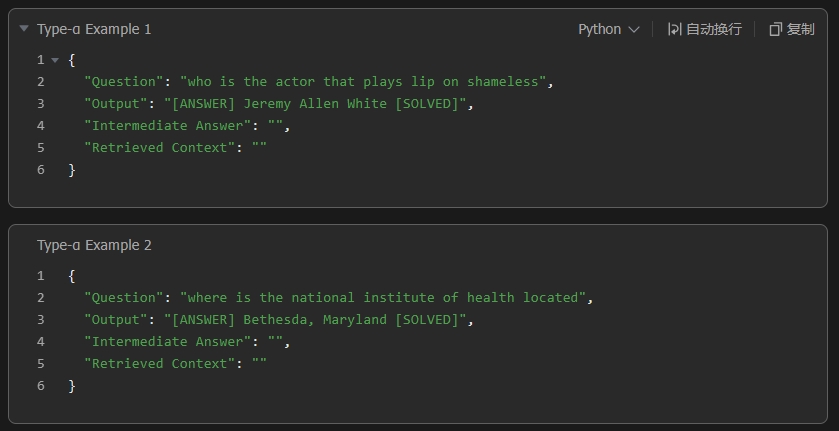}
    \caption{\textbf{Type-\boldmath$\alpha$}: Samples that are directly answerable using the model's internal knowledge alone, where the model emits \texttt{[ANSWER]} followed by the answer and concludes with \texttt{[SOLVED]} without any retrieval.}
    \label{pic.type1}
\end{figure}

\begin{figure}[h]
    \centering
    \includegraphics[width=1.0\linewidth]{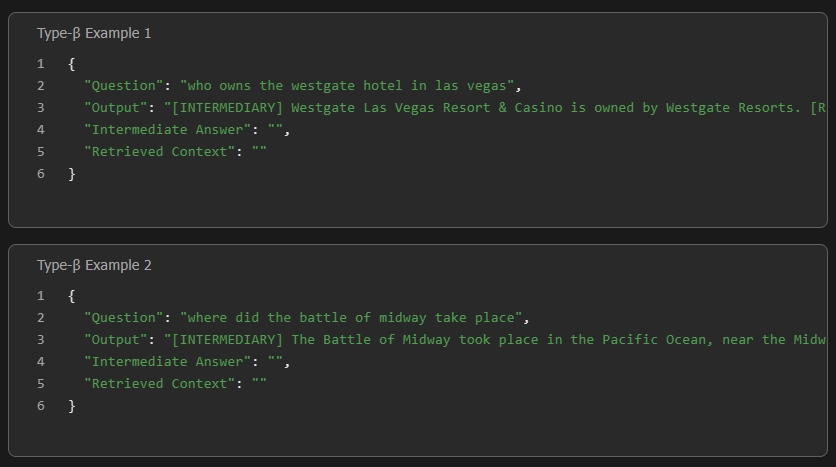}
    \caption{\textbf{Type-\boldmath$\beta$}: Samples where the model produces partial or noisy answers containing the gold answer but lacking clarity, encouraging it to recognize uncertainty and issue a retrieval request through \texttt{[INTERMEDIARY]} followed by \texttt{[RETRIEVE]}.}
    \label{pic.type1}
\end{figure}

\begin{figure}[h]
    \centering
    \includegraphics[width=1.0\linewidth]{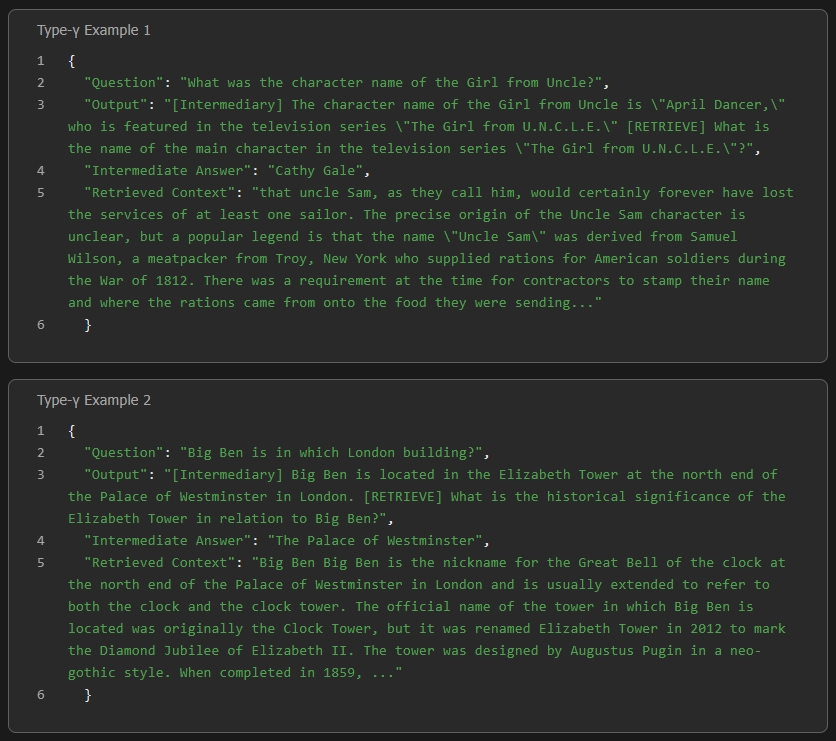}
    \caption{\textbf{Type-\boldmath$\gamma$}: Multi-hop or complex knowledge samples where both parametric memory and basic retrieval fail, requiring the model to iteratively construct better sub-questions and reason across multiple planning steps before concluding with \texttt{[ANSWER]} and \texttt{[SOLVED]}.}
    \label{pic.type1}
\end{figure}

\begin{figure}[h]
    \centering
    \includegraphics[width=1.0\linewidth]{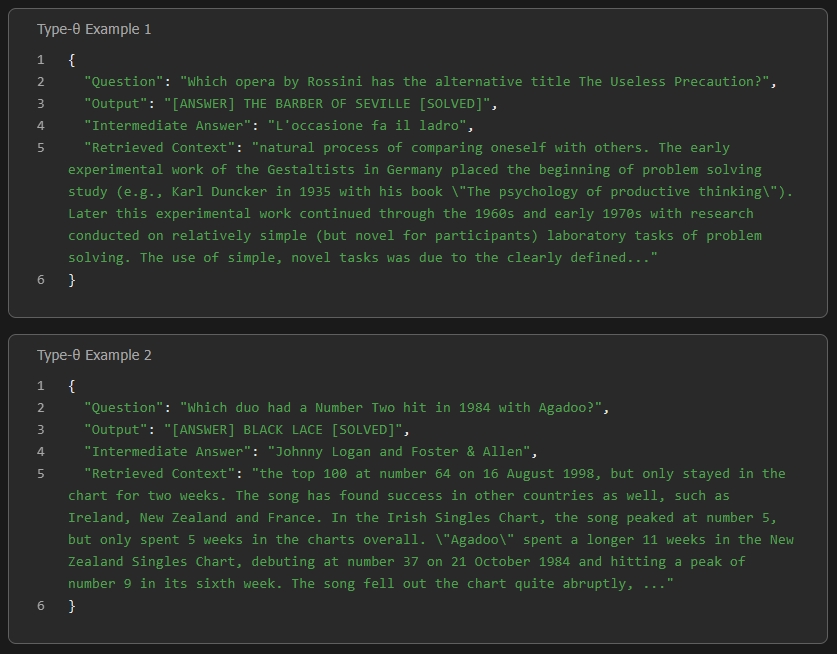}
    \caption{\textbf{Type-\boldmath$\theta$}: Cases where the retrieved content covers the answer superficially but demands further reasoning and content compression, training the model to selectively synthesize key information via \texttt{[INTERMEDIARY]} $\rightarrow$ \texttt{[RETRIEVE]} $\rightarrow$ \texttt{[ANSWER]} transitions.}
    \label{pic.type1}
\end{figure}

\section{Various Control Tokens Settings}\label{app:token}

In the preliminary stage of GRIP, we explored multiple strategies for defining control tokens used to guide the decoding process. Specifically, we experimented with the following approaches: 

\begin{itemize}
    \item Using natural-language-style tokens such as \texttt{[ANSWER]} or \texttt{[RETRIEVE]} directly as special control tokens;
    
    \item Replacing control tokens with reserved but unused tokens from the pretrained model vocabulary, such as \texttt{[ANSWER]} replaced by \texttt{<|reserved\_special\_token\_51|>};
    
    \item Introducing four newly defined tokens into the tokenizer vocabulary and initializing them with embeddings of semantically related words. For example, \texttt{[RETRIEVE]} is initialized with the embedding of the word \texttt{retrieve}.
\end{itemize}

Empirical results showed that these strategies resulted in only marginal differences in overall model performance. To reduce implementation complexity and ensure compatibility with standard decoding processes, we selected the first approach for all subsequent experiments.

\section{Ablation on the Reward Metric}
\label{app:reward_metric_ablation}

\begin{table}[h]
\small
\centering
\setlength{\tabcolsep}{2.2mm}
\renewcommand{\arraystretch}{1.12}
\begin{tabular}{c|c|c}
\toprule[1.5pt]
\textbf{Reward metric for $r_{\text{ans}}$} & \textbf{Avg.Score} & \textbf{Avg.Count}\\
\midrule
BLEU (default) & \textbf{41.0} & \textbf{1.24}\\
ROUGE & 40.5 & \textbf{1.24}\\
F1    & 40.8 & 1.25\\
EM    & 40.6 & \textbf{1.24}\\
\bottomrule[1.5pt]
\end{tabular}
\caption{Reward-metric ablation for RL: we only change the overlap metric used in the answer-fidelity reward $r_{\text{ans}}$; all other settings are fixed.}
\label{tab:rl_reward_metric_ablation}
\end{table}

We replace the answer-fidelity reward $r_{\text{ans}}$ with BLEU, ROUGE, token-F1, or EM while keeping the RL pipeline unchanged.
As shown in Table~\ref{tab:rl_reward_metric_ablation}, performance is highly consistent across metrics (within 0.5 Avg.Score), and the average number of retrieval calls is nearly unchanged.
This suggests that the RL stage is not sensitive to a specific overlap definition and primarily provides fine-grained shaping of answer quality.
The near-identical Avg.Count indicates that the control reward dominates the retrieval policy, while $r_{\text{ans}}$ mainly affects answer wording rather than when to retrieve.
We use BLEU by default, as it offers a smoother graded signal than strict EM while achieving the best Avg.Score under essentially the same retrieval budget.

\section{Behavioral Shift after Rule-based RL}\label{app:shift}

\begin{figure}[h]
    \centering
    \includegraphics[width=1.0\linewidth]{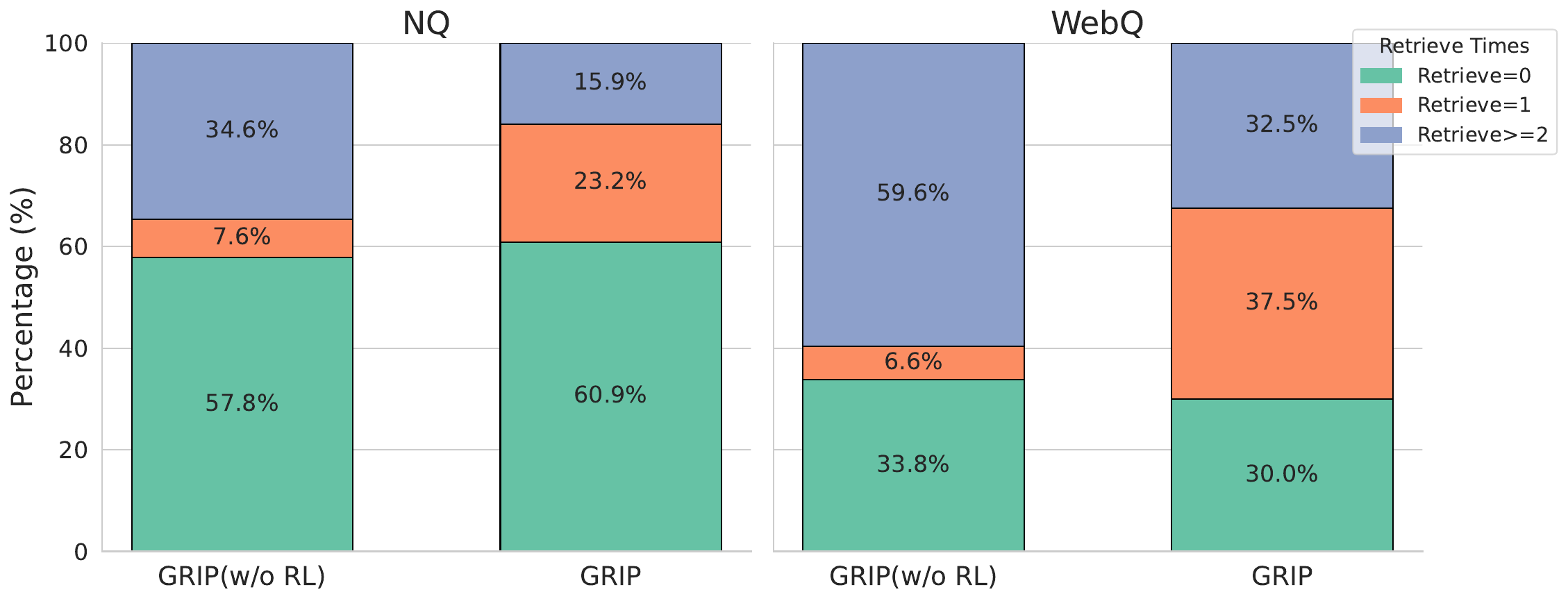}
    \caption{Comparison of retrieval behavior distributions before and after reinforcement learning on NQ and WebQ.}
    \label{pic.retrieve_distribution}
\end{figure}

We analyze GRIP’s retrieval behavior before and after reinforcement learning (GRIP (w/o RL) vs. GRIP) on two representative datasets: NQ and WebQ. As shown in Figure~\ref{pic.retrieve_distribution}, Rule-based RL significantly reduces unnecessary second retrieval steps, e.g., the proportion of Retrieve=2 on WebQ drops from 59.6\% to 32.5\%, while Retrieve=1 rises markedly from 6.6\% to 37.5\%. This shift indicates that GRIP learns to make earlier stopping decisions once sufficient context is available, thereby avoiding over-retrieval. On NQ, where many questions can be answered directly, GRIP maintains a high Retrieve=0 rate (from 57.8\% to 60.9\%), confirming its ability to preserve efficient strategies on easier tasks. Overall, these results highlight GRIP's capacity to adapt its retrieval depth based on the complexity of individual questions, demonstrating more interpretable and context-aware control over retrieval behaviors.

Overall, RL mainly converts many ``Retrieve=2'' cases into ``Retrieve=1'' rather than increasing ``Retrieve=0'', suggesting it improves early stopping once sufficient evidence is obtained.

\section{Case Studies of GRIP’s Retrieval Planning Behavior}
To better illustrate the internal behavior of GRIP during retrieval-augmented generation, we present a series of qualitative case studies that highlight how the model dynamically plans, regulates, and adapts its retrieval strategy. These examples cover representative scenarios including retrieval under information insufficiency, multi-step query refinement, and generation under explicit control signals. Each case demonstrates a different aspect of GRIP’s token-level decision-making process, providing deeper insights into its reasoning trajectory, interpretability, and interaction between generation and retrieval.

\subsection*{Case 1: Fallback to Internal Knowledge When GRIP Cannot Answer}
To understand how GRIP behaves when external information is insufficient or retrieval is delayed, we examine a case where the model fails to answer the original question. Instead of hallucinating a confident response, GRIP generates partial reasoning steps based on its internal knowledge, revealing its awareness of information gaps. This demonstrates GRIP's ability to initiate cautious reasoning before retrieval is triggered, contrasting with methods like DRAGIN that may prematurely commit to incomplete answers.

\begin{figure}[h]
    \centering
    \includegraphics[width=1.0\linewidth]{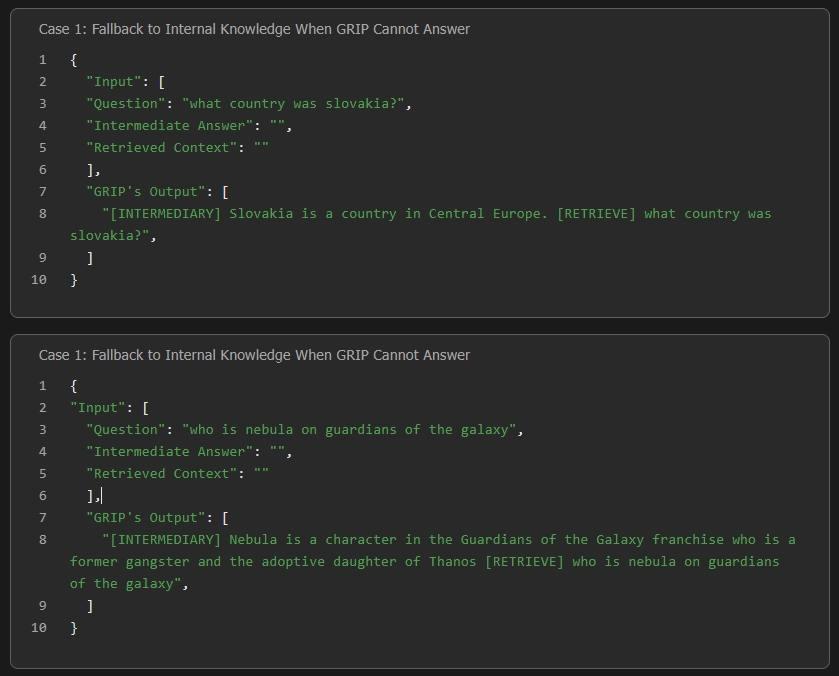}
    \caption{Example of GRIP falling back to internal knowledge and deferring retrieval when no context is available.}
    \label{fig:case1}
\end{figure}

In both examples shown in Figure~\ref{fig:case1}, GRIP is presented with questions for which no retrieved evidence is initially available. Rather than producing a low-confidence final answer, GRIP emits an \texttt{[INTERMEDIARY]} token followed by an internally composed partial response grounded in its parametric knowledge. For instance, when asked ``what country was Slovakia?'' or ``who is Nebula on Guardians of the Galaxy'', GRIP provides factual context (e.g., ``Slovakia is a country in Central Europe'' or ``Nebula is a character in the Guardians franchise...''), before emitting \texttt{[RETRIEVE]} along with a reformulated sub-question. This pattern reflects GRIP's ability to recognize uncertainty, enabling it to avoid hallucination while still making forward progress in reasoning. Such fallback behavior illustrates a core strength of GRIP: its capacity to delay final commitment until sufficient evidence is gathered.

\subsection*{Case 2: GRIP Refines Queries Through Iterative Retrieval Planning}
We present a multi-step example in which GRIP issues several \texttt{[RETRIEVE]} signals, each followed by a newly generated query. Initially, the retrieved content is insufficient to support a final answer. However, GRIP leverages its intermediate outputs to progressively formulate more focused and precise new queries. This behavior illustrates the model’s capacity for retrieval planning: dynamically adjusting its information-seeking behavior based on context evolution and accumulated evidence.

\begin{figure}[h]
    \centering
    \includegraphics[width=1.0\linewidth]{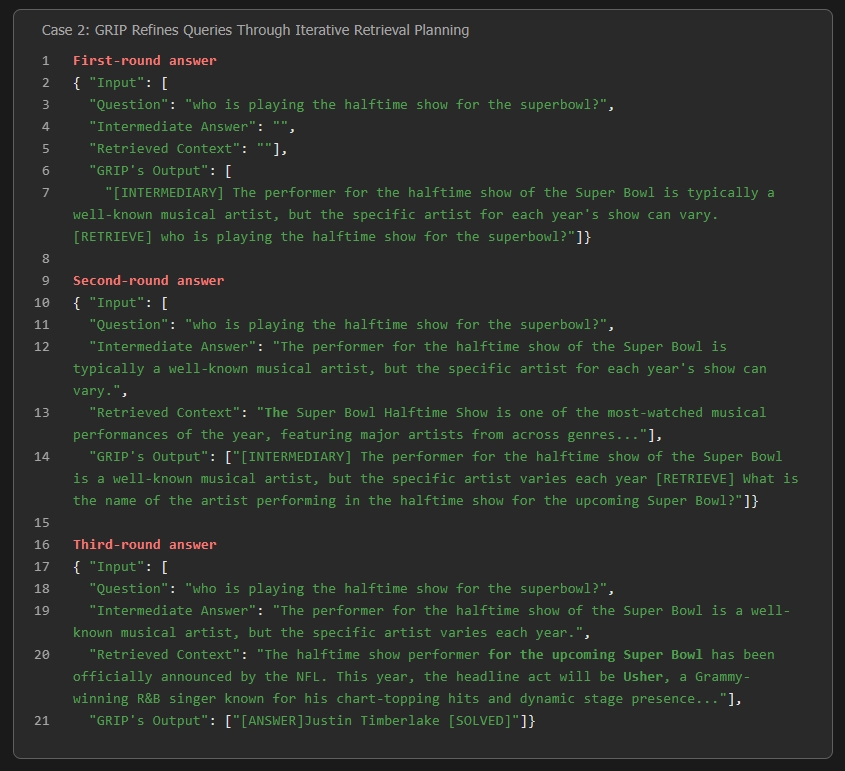}
\caption{Example of GRIP refining its retrieval query over multiple rounds to obtain a better answer.}
    \label{fig:case2}
\end{figure}

In this case (Figure~\ref{fig:case2}), we observe that the intermediate answer produced in the first and second rounds is nearly identical, indicating that the initial retrieval step failed to surface useful content for answering the original question. Despite this, GRIP does not prematurely conclude. Instead, it formulates a more specific and context-aware query in the second round: ``What is the name of the artist performing in the halftime show for the upcoming Super Bowl?’’ This refined question enables the retriever to access a higher-quality passage containing the correct entity. In the third round, GRIP successfully emits \texttt{[ANSWER]} and \texttt{[SOLVED]}, demonstrating that effective retrieval often hinges not on initial coverage but on the model's ability to iteratively reformulate sub-questions until the right evidence is found. This case showcases GRIP's retrieval planning mechanism in action, where small linguistic changes in the generated query lead to significant downstream improvements in answerability.

\subsection*{Case 3: GRIP’s Summarization Behavior by Answering with \texttt{[ANSWER]}}

In this case, we investigate how GRIP responds when explicitly instructed to finalize an answer via the \texttt{[ANSWER]} control token. Despite not having retrieved all necessary evidence, the model integrates its current reasoning trajectory and partial observations to generate a coherent summary response. This highlights GRIP’s ability to adapt to control signals and perform best-effort answering under constrained or premature termination, showcasing a key advantage of token-level behavioral regulation.

\begin{figure}[h]
    \centering
    \includegraphics[width=1.0\linewidth]{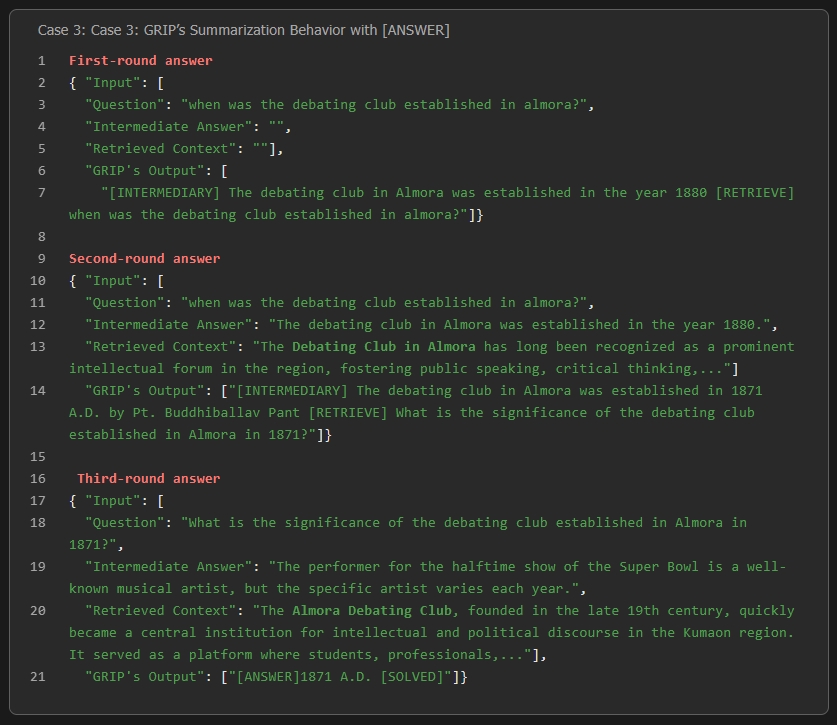}
\caption{Example of GRIP refining, reformulating, and compressing retrieved knowledge into a final answer using \texttt{[ANSWER]}.}
    \label{fig:case3}
\end{figure}

Figure~\ref{fig:case3} illustrates a chained reasoning trajectory where GRIP first generates an incorrect intermediate answer based on parametric memory (\texttt{1880}), then revises it after retrieving a more accurate year (\texttt{1871}) from external content. Rather than concluding immediately, the model goes one step further: it reformulates the question to explore the historical significance of the event, triggering another retrieval round. Finally, GRIP compresses the relevant passage into a concise factual response (\texttt{1871 A.D.}) using the \texttt{[ANSWER]} and \texttt{[SOLVED]} tokens. This case demonstrates GRIP's ability to integrate evidence across turns, refine its beliefs based on new context, and summarize multi-step reasoning into a final answer—without hallucinating or skipping steps. It also showcases how retrieval can reshape the reasoning trajectory itself, not just inform it.

\section{Failure Analysis }
This section presents selected failure cases of GRIP to better understand its limitations in retrieval-augmented generation. While GRIP exhibits strong performance across benchmarks, we observe occasional failure patterns that reveal important areas for improvement. Specifically, we analyze representative cases involving incorrect final answers and redundant retrieval behaviors.

\subsection*{Case 1: Incorrect Final Answer}
In some cases, GRIP successfully retrieves relevant evidence but fails to generate the correct final answer. This often occurs when the retrieved passages contain implicit or multi-faceted information that requires deeper logical inference. These failures highlight limitations in GRIP’s reasoning capability under answer ambiguity or insufficient summarization, even when the retrieval component functions as intended.

\subsection*{Case 2: Redundant Retrieval Steps}
GRIP’s token-level control enables flexible retrieval depth, but in certain examples, the model triggers unnecessary retrievals that do not improve or even dilute answer quality. Such cases typically occur when the model fails to recognize that sufficient information is already available, resulting in redundant sub-queries and noisy context accumulation. These examples illustrate the challenges of balancing retrieval sensitivity and reasoning sufficiency, especially in high-recall settings.

\section{Latent State Visualization of Control Decisions}

\begin{figure}[h]
\centering
\includegraphics[width=0.8\linewidth]{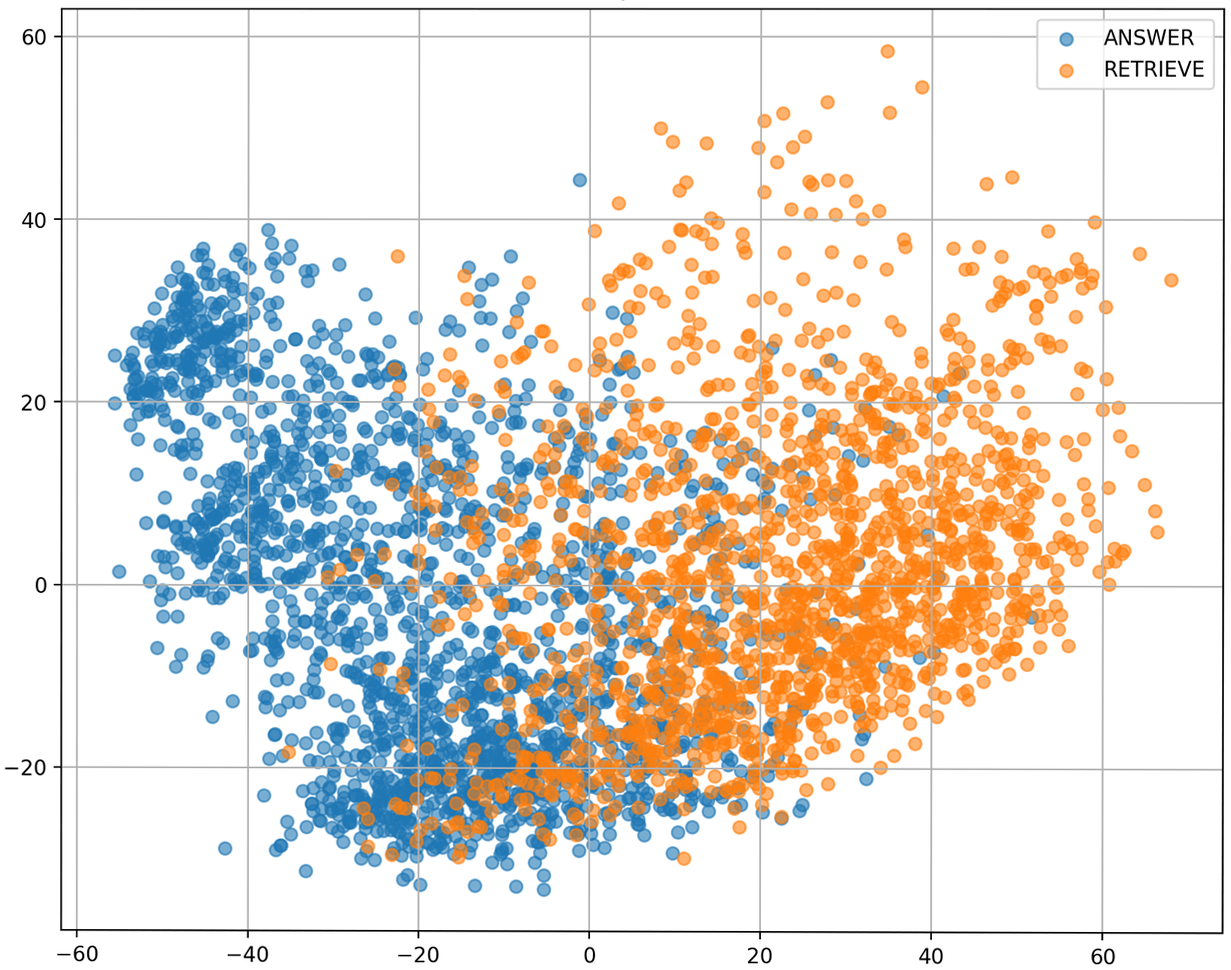}
\caption{PCA visualization of the model’s hidden states before emitting \texttt{[ANSWER]} (blue) and \texttt{[RETRIEVE]} (orange) on the NQ dataset. Each point corresponds to the hidden representation of the last token in the input prompt. The distinct clusters indicate that the model encodes separable internal states for different control decisions.}
\label{fig:pca_nq_control}
\end{figure}

To better understand the model's internal decision dynamics, we visualize the hidden representations associated with the two control tokens: \texttt{[ANSWER]} and \texttt{[RETRIEVE]}. For each sample, we extract the hidden state of the last token in the input prompt, immediately before the model emits a control token. These hidden states represent the model's contextual understanding just prior to triggering a retrieval or producing an answer. We then apply principal component analysis (PCA) to reduce the hidden states to two dimensions and plot the results in Figure~\ref{fig:pca_nq_control}.

The visualization reveals a clear separation between the two control behaviors in the latent space. Samples that lead to \texttt{[RETRIEVE]} tend to cluster in a distinct region from those that lead to \texttt{[ANSWER]}. This suggests that the model has learned to encode different latent states for retrieval versus answer emission, reinforcing the effectiveness of control-token-based behavioral modulation.

\section{Prompts Used in Structured Data Generation}\label{app:prompt}

As part of our data generation pipeline, we employ GPT-based prompting to simulate how a model might revise its internal reasoning and generate improved follow-up queries. Given the original question, a partial (intermediary) answer, and a set of retrieved documents, the prompt instructs the model to refine the known fact and formulate a more targeted query that increases the likelihood of successful retrieval. This process is used to construct structured training data that supports GRIP’s ability to perform iterative retrieval planning in a self-supervised fashion.

\begin{figure}[h]
\centering
\includegraphics[width=0.99\linewidth]{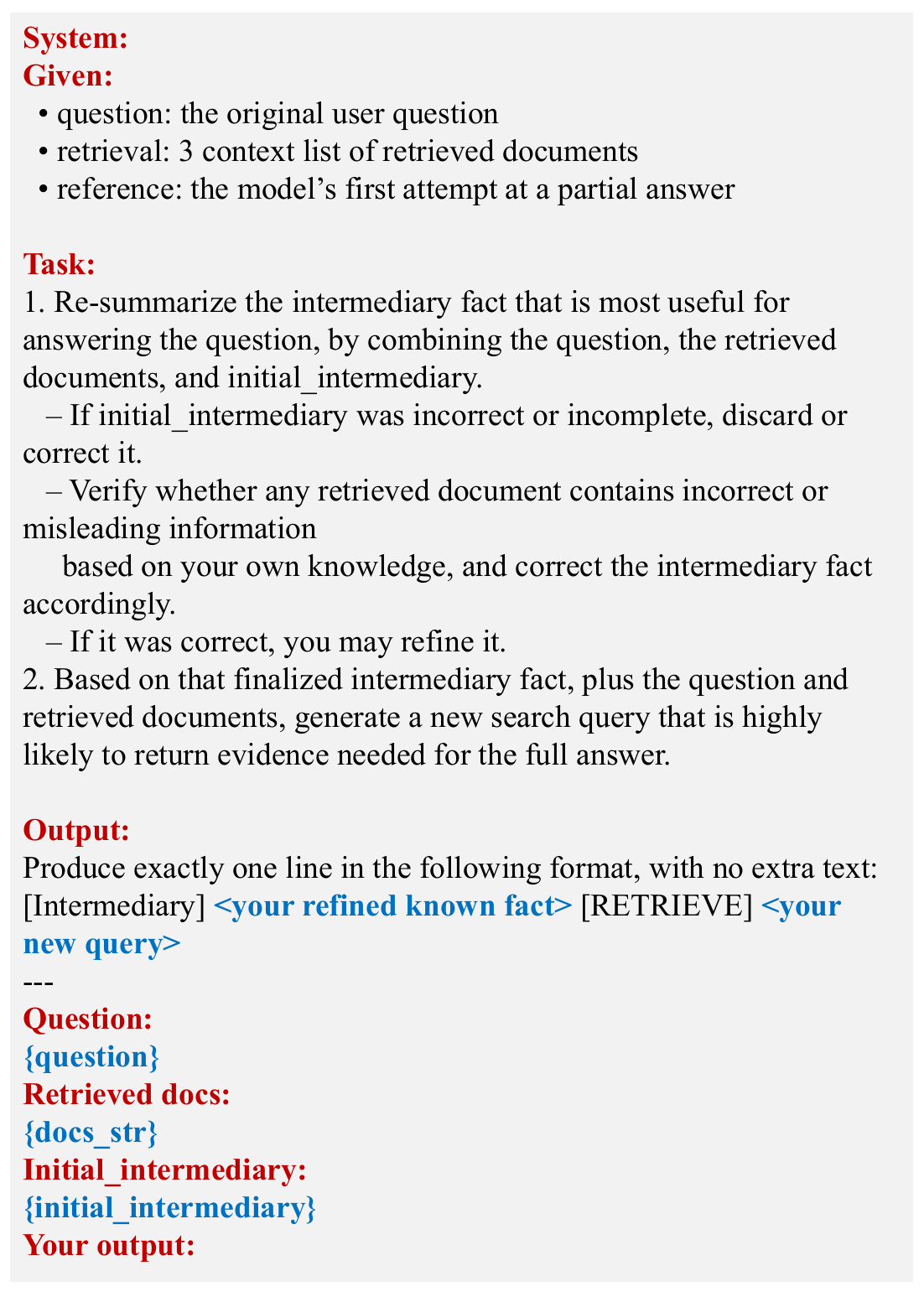}
\caption{Prompt used for data generation with GPT-4o-mini.}
\label{fig:prompt}
\end{figure}

\section{Retriever Ablation}
\label{app:retriever_ablation}

\begin{table}[]
\small
\centering
\setlength{\tabcolsep}{1.2mm}
\renewcommand{\arraystretch}{1.15}
\begin{tabular}{l|ccc}
\toprule[1.5pt]
\textbf{Method} & \textbf{BM25} & \textbf{DPR} & \textbf{Hybrid} \\
\midrule[1.5pt]
Single-RAG (training-free) & 30.8 & 31.9 & 33.7 \\
RobustRAG (training-based) & 33.2 & 32.1 & 33.8 \\
GRIP              & \textbf{41.0} & 39.2 & \textbf{41.0} \\
\bottomrule[1.5pt]
\end{tabular}
\caption{Retriever ablation on the main evaluation suite. We replace the default BM25 retriever with a dense DPR retriever and a BM25+DPR hybrid retriever (denoted \textbf{Hybrid}), while keeping the backbone (LLaMA3-8B), corpus (Wikipedia), and top-$k$ passages ($k{=}3$) fixed. Single-RAG and RobustRAG are representative training-free and training-based baselines from the main table, respectively.}
\label{tab:retriever_ablation}
\end{table}

To assess how sensitive our conclusions are to the choice of retriever, we evaluate GRIP and two representative baselines under three retrieval backends: sparse BM25, dense DPR, and a BM25+DPR hybrid retriever. We select Single-RAG as a representative training-free method and RobustRAG as a representative training-based method. For each retriever, we keep all other factors unchanged, including the LLaMA3-8B backbone, the Wikipedia corpus, and the number of retrieved passages ($k{=}3$).

Table~\ref{tab:retriever_ablation} shows that hybrid retrieval consistently improves performance for both Single-RAG and RobustRAG, indicating that combining sparse and dense signals can provide higher-quality evidence than either retriever alone. For GRIP, DPR alone is slightly worse than BM25, while the hybrid retriever matches the best result. Overall, GRIP remains substantially stronger than both baseline families across all retriever choices, suggesting that its gains stem from the proposed token-level planning and training recipe rather than a particular retriever implementation. In the main paper we adopt BM25 as the default retriever for efficiency and controlled comparison, and report these additional retriever variants here for completeness.

\section{Pseudocode of GRIP Decoding with Self-Triggered Information Planning}\label{app:code}

The following pseudocode outlines the decoding procedure of GRIP, which integrates retrieval control into autoregressive generation through self-triggered information planning. It illustrates how the model emits control tokens to decide when to retrieve, generate sub-questions, and finalize answers during the decoding process.

\begin{algorithm}
\caption{GRIP Decoding with Self-Triggered Information Planning}
\begin{algorithmic}[1] 
\Statex \textbf{Input:} query $x$, model $\pi_\theta$, retriever $\mathcal{R}$, max steps $B$
\Statex \textbf{Output:} final answer $y$
\State Initialize $y \gets \emptyset$, $m \gets \emptyset$, $b \gets 0$
\While{$b < B$}
  \State Sample $t \sim \pi_\theta(x, m, y)$
  \State Append $t$ to $y$
  \If{$t =$ [ANSWER]}
    \State Generate final answer until [SOLVED]
    \State \Return $y$
  \ElsIf{$t =$ [INTERMEDIARY]}
    \State Generate partial answer $a$, then generate next token $t'$
    \If{$t' =$ [RETRIEVE]}
      \State Generate new query $q$
      \State Retrieve $c \gets \mathcal{R}(q)$
      \State $m \gets m \cup \{c\}$; $b \gets b + 1$
    \Else
      \State Append fallback message
    \EndIf
  \ElsIf{$t =$ [SOLVED]}
    \State \Return $y$
  \EndIf
\EndWhile
\State \Return $y$
\end{algorithmic}
\end{algorithm}

\section{Retrieval-Count Distributions under Budget Control}
\label{app:budget}

\begin{figure*}[h]
\centering
\includegraphics[width=0.99\linewidth]{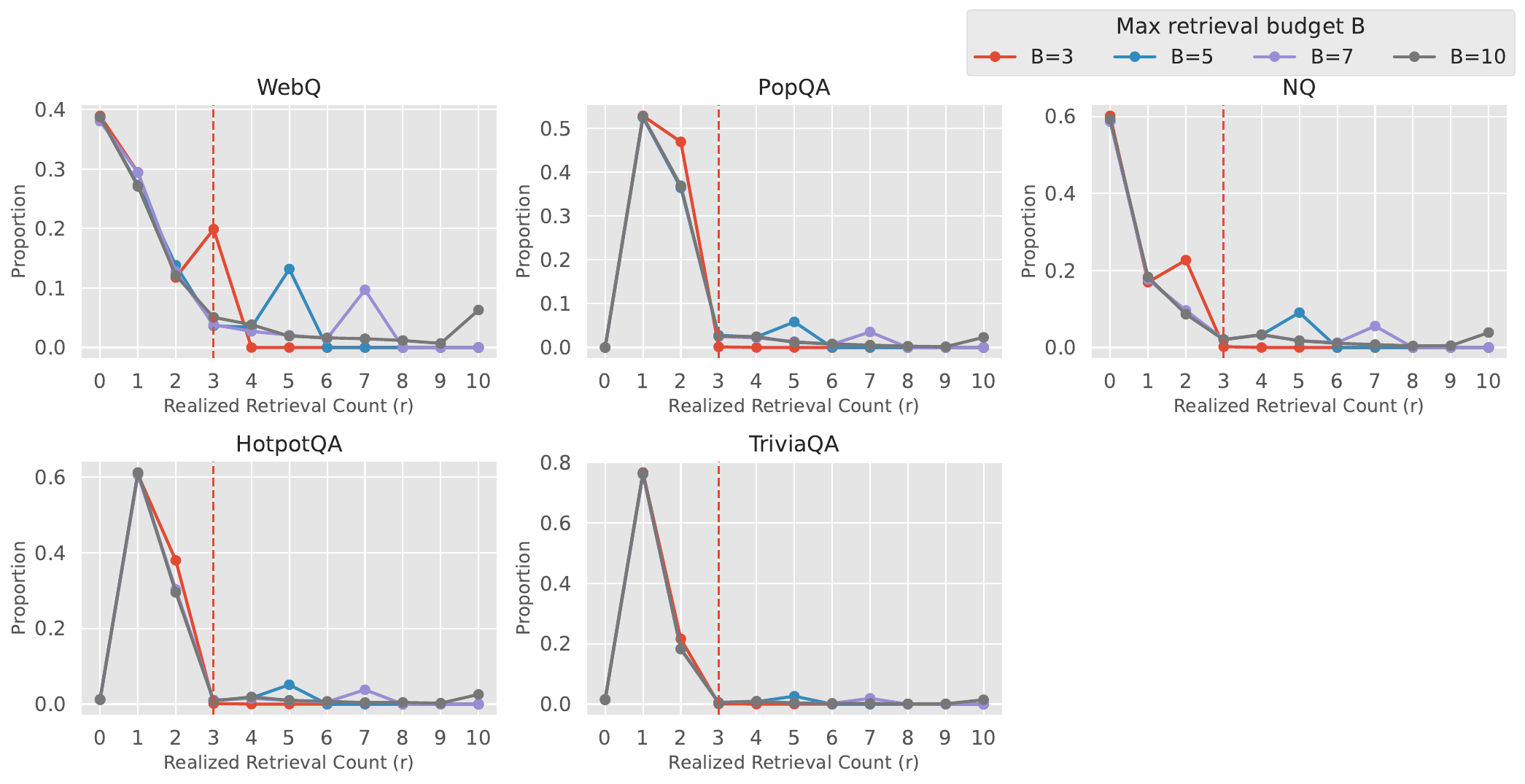}
\caption{Realized retrieval-count distributions of GRIP under different maximum retrieval budgets $B$.
Each subplot corresponds to one benchmark; curves show the fraction of examples that trigger $r$ retrieval calls at inference time.
The dashed vertical line marks $r{=}3$, the maximum retrieval depth observed in our training trajectories.
As $B$ increases, GRIP places more probability mass beyond $r{>}3$, indicating extrapolation beyond the training depth, while the distribution remains concentrated at small $r$, suggesting selective and task-aware retrieval rather than mechanically consuming the available budget.}
\label{fig:retrieval_count_dist}
\end{figure*}

We vary the inference-time maximum retrieval budget $B \in \{3,5,7,10\}$ and record the realized number of retrieval calls $r$ per example.
Figure~\ref{fig:retrieval_count_dist} reports the empirical distribution of $r$ on each benchmark.
The dashed vertical line highlights $r{=}3$, which matches the maximum retrieval depth used in our constructed training trajectories. Across all datasets, GRIP does not saturate the budget: even when $B$ is large, most probability mass remains on small $r$ (typically $r \le 2$), indicating that retrieval is triggered selectively when the decoding trajectory suggests an information gap, instead of being executed as a fixed schedule.

Importantly, the figure also demonstrates \emph{extrapolative generalization} beyond the training depth.
Our training trajectories are constructed with at most three retrieval steps, yet increasing $B$ at test time yields non-trivial mass at $r{>}3$.
At $B{=}10$, the fraction of examples with $r{>}3$ reaches 17.0\% on WebQ, 11.7\% on NQ, 7.8\% on PopQA, 7.3\% on HotpotQA, and 3.4\% on TriviaQA.
This shows that GRIP learns a transferable decision principle of \emph{when additional evidence is needed}, rather than overfitting to a fixed maximum-step pattern seen during training.

Finally, the extent of extrapolation is task-dependent.
Datasets that are more evidence-uncertain exhibit heavier tails beyond $r{=}3$ (e.g., WebQ and NQ), whereas factoid-style benchmarks concentrate more mass at $r{=}1$ or $r{=}2$ and rarely exceed the training depth (e.g., TriviaQA).
Overall, budget control provides a practical mechanism to trade retrieval cost for accuracy while preserving GRIP's adaptive behavior beyond the retrieval depth seen in training.

\section{DAPO Configuration for the RL Stage}
\label{app:dapo_config}

We adopt the official DAPO recipe from the \textsc{VERL} codebase and keep its core algorithmic configuration unchanged; our modifications are limited to the data configuration and our task-specific reward definitions described in \S\ref{sec:rl_reward}.%
\footnote{\url{https://github.com/volcengine/verl/blob/main/recipe/dapo/run_dapo_qwen2.5_32b.sh}}
In particular, we enable CLIP-HIGHER, token-level policy gradient loss, dynamic sampling, and soft overlong reward shaping, and disable KL regularization (both KL-in-reward and KL loss).

\begin{table*}[t]
\centering
\small
\setlength{\tabcolsep}{3.8mm}
\renewcommand{\arraystretch}{1.10}
\begin{tabular}{l|c|l}
\toprule[1.5pt]
\textbf{Component} & \textbf{Status} & \textbf{Key settings} \\
\midrule
CLIP-HIGHER & on &
clip\_ratio\_low=0.2,\; clip\_ratio\_high=0.28,\; clip\_ratio\_c=10.0 \\
Token-level PG loss & on &
loss\_agg\_mode=\texttt{token-mean} \\
Dynamic sampling & on &
enabled (default DAPO setting) \\
Soft overlong shaping & on &
overlong\_buffer.len=8192,\; penalty\_factor=1.0 \\
\midrule
KL in reward & off &
use\_kl\_in\_reward=False,\; kl\_coef=0.0 \\
KL loss & off &
use\_kl\_loss=False,\; kl\_loss\_coef=0.0 \\
\bottomrule[1.5pt]
\end{tabular}
\caption{Key DAPO components used in our RL stage. All unspecified hyperparameters follow the referenced DAPO recipe.}
\label{tab:dapo_components}
\end{table*}

Soft overlong punishment is enabled and activates when the response length exceeds \texttt{overlong\_buffer.len}. Dynamic sampling is enabled as in the default recipe to encourage advantage diversity during RL training.

\section{Results with Qwen2.5-7B}
\label{app:qwen}

To verify that our findings are not specific to a single backbone, we additionally evaluate GRIP and the reproduced baselines using \textbf{Qwen2.5-7B} under the same evaluation protocol. In particular, we keep the external corpus (Wikipedia), retriever (BM25), and top-$k$ passages ($k{=}3$) unchanged, and apply the same decoding and scoring procedure as in the main experiments. 

The results on Qwen2.5-7B exhibit the \emph{same trend} as those on LLaMA3-8B: GRIP consistently outperforms both training-free and training-based RAG baselines across datasets, while maintaining strong retrieval efficiency through token-level planning. The relative ranking among baseline methods is also stable, indicating that the gains of GRIP stem from the proposed retrieval-as-generation formulation and scenario-typed supervision rather than backbone-specific effects. We further observe that the RL stage provides an additional but modest improvement on Qwen2.5-7B, consistent with our main findings that RL primarily refines retrieval behavior and stabilizes termination decisions.

\begin{table*}[t]
\centering
\small
\setlength{\tabcolsep}{3.5pt}
\renewcommand{\arraystretch}{1.15}
\resizebox{\textwidth}{!}{
\begin{tabular}{lcccccccccccccccc}
\toprule[1.5pt]
& \multicolumn{3}{c}{HotpotQA} & \multicolumn{3}{c}{PopQA} & \multicolumn{3}{c}{NQ} & \multicolumn{3}{c}{WebQ} & \multicolumn{3}{c}{TriviaQA} & \multirow{2}{*}{Avg.\ Score} \\
\cmidrule(lr){2-4} \cmidrule(lr){5-7} \cmidrule(lr){8-10} \cmidrule(lr){11-13} \cmidrule(lr){14-16}
& EM & ROUGE & F1 & EM & ROUGE & F1 & EM & ROUGE & F1 & EM & ROUGE & F1 & EM & ROUGE & F1 &  \\
\toprule[1.5pt]
\multicolumn{17}{l}{\textbf{Qwen-3-4B-Instruct}} \\\toprule[1.5pt]
Instruct        &  8.0 & 13.6 & 16.5 &  5.3 &  8.0 & 10.3 &  6.2 & 11.3 & 13.7 &  7.7 & 16.3 & 20.0 & 22.1 & 24.1 & 30.1 & 14.2 \\
Single-RAG      & 18.4 & 24.6 & 28.8 & 18.3 & 21.7 & 25.5 & 12.9 & 18.9 & 21.9 &  8.7 & 16.7 & 20.2 & 36.0 & 38.0 & 46.0 & 23.8 \\
Robust-RAG      & 27.3 & 30.8 & 36.7 & 27.3 & 28.2 & 33.2 & 20.9 & 23.8 & 28.1 & 19.8 & 24.5 & 28.7 & 47.7 & 45.2 & 55.1 & 31.8 \\
R1-Searcher     & 18.2 & 21.2 & 25.4 & 12.5 & 14.4 & 18.0 & 11.8 & 16.8 & 20.1 & 12.9 & 21.6 & 25.4 & 36.7 & 35.1 & 43.5 & 22.3 \\
GRIP   & 31.9 & 39.7 & 46.2 & 27.8 & 30.9 & 36.6 & 21.1 & 28.6 & 33.2 & 16.8 & 27.5 & 32.3 & 49.9 & 48.2 & 59.1 & 35.3 \\
\toprule[1.5pt]
\multicolumn{17}{l}{\textbf{Qwen2.5-7B-Instruct}} \\\toprule[1.5pt]
Instruct        & 18.3 & 21.5 & 26.0 & 11.5 & 13.0 & 16.3 & 11.8 & 16.7 & 20.0 & 14.5 & 22.6 & 26.4 & 37.1 & 36.0 & 44.4 & 22.4 \\
Single-RAG      & 30.6 & 34.9 & 40.9 & 23.6 & 26.6 & 31.1 & 18.9 & 24.6 & 28.3 & 15.3 & 22.6 & 26.3 & 45.8 & 46.1 & 55.7 & 31.4 \\
Robust-RAG      & 26.1 & 30.8 & 35.5 & 27.5 & 28.6 & 33.5 & 20.5 & 23.8 & 27.8 & 21.1 & 25.8 & 30.0 & 48.7 & 46.7 & 56.7 & 32.2 \\
R1-Searcher     & 20.2 & 24.3 & 28.8 & 11.6 & 13.8 & 17.0 & 14.3 & 19.2 & 23.0 & 19.3 & 26.7 & 30.7 & 40.1 & 39.4 & 48.3 & 25.1 \\
GRIP   & 32.0 & 37.6 & 44.6 & 28.7 & 30.9 & 36.3 & 21.2 & 27.7 & 32.5 & 19.3 & 28.3 & 33.2 & 51.9 & 49.6 & 60.5 & 35.6 \\
\toprule[1.5pt]
\end{tabular}
}
\caption{Main results on five QA benchmarks under two backbones. We report EM, ROUGE, F1, and Avg.\ Score.}
\label{tab:main_results_qwen}
\end{table*}

\end{document}